\documentclass[twocolumn]{IEEEtran}

\pdfoutput=1 

\usepackage{amsmath}
\usepackage{amssymb}
\usepackage{graphicx}
\usepackage{subfigure}
\usepackage{cite}
\usepackage{multirow}

\DeclareMathOperator*{\argmax}{arg\,max}
\DeclareMathOperator*{\argmin}{arg\,min}

\DeclareGraphicsExtensions{%
.jpg,.JPG,%
}

\usepackage[hidelinks,pdfstartview=Fit]{hyperref}
\usepackage{color}
\definecolor{magenta}{rgb}{.5,.25,.5}

\newcommand{\Hdeletetext}[1]{{\color{blue}x}}

\newcommand{\Cdeletetext}[1]{{\color{magenta}x}}

\begin{document}

\title{3D Display Calibration by Visual Pattern Analysis}

\author{Hyoseok Hwang, \quad Hyun Sung Chang, 
\quad Dongkyung Nam, \quad In So Kweon
\thanks{
Manuscript submitted January 1, 2016.
This work extends the previous work
presented at a conference \cite{hwang2015lfcalib}.
}
\thanks{
Hyoseok Hwang and In So Kweon are
with the Department of Electrical Engineering,
Korea Advanced Institute of Science and Technology (KAIST),
291 Daehak-ro, Yuseong-gu, Daejon 34141, South Korea
(e-mail: \{hyoseok.hwang,\,iskweon\}@kaist.ac.kr).
H.~Hwang is also affiliated with
Samsung Advanced Institute of Technology (SAIT),
130 Samsung-ro, Yeongtong-gu, Suwon 16678, South Korea.
}
\thanks{
Hyun Sung Chang and Dongkyung Nam are with SAIT 
(e-mail: \{hyun.s.chang,\,max.nam\}@samsung.com).
}
}

\maketitle

\begin{abstract}
Nearly all 3D displays need calibration for correct rendering.
More often than not, the optical elements
in a 3D display are misaligned from the designed parameter setting.
As a result, 3D magic does not perform well as intended.
The observed images tend to get distorted.

In this paper, we propose a novel display calibration method
to fix the situation.
In our method, a pattern image is displayed on the panel and a camera
takes its pictures twice at different positions.
Then, based on a quantitative model,
we extract all display parameters
(i.e., pitch, slanted angle, gap or thickness, offset)
from the observed patterns in the captured images.
For high accuracy and robustness,
our method analyzes the patterns mostly in frequency domain.
We conduct two types of experiments for validation;
one with optical simulation for quantitative results
and the other with real-life displays
for qualitative assessment.
Experimental results demonstrate that
our method is quite accurate,
about a half order of magnitude higher
than prior work;  
is efficient, spending less than 2\,s for computation;
and is robust to noise, working well in the SNR regime
as low as 6\,dB.
\end{abstract}


\section{Introduction}

\PARstart{D}{uring} the past decade, we have witnessed
rapid advancement in 3D display technology.
Diverse approaches have attempted
to realize 3D scenes on a flat panel display. 
Among them, a multi\-view auto\-stereoscopic display provides,
at a low cost, immersive 3D environments
to multiple users without requirement of wearing special glasses.
An auto\-stereoscopic display uses optical elements
such as lenticular lenses or parallax barriers
to direct each ray from the pixels
toward the intended view point \cite{dodgson2013optical,urey2011state}.
Although the lenses and barriers work differently,
i.e., lenses by refracting the rays
and barriers by blocking irrelevant rays,
they commonly fulfill the same principle
that different pixels must be visible
in different eye positions.
If the left eye and the right eye are located
in different viewing zones,
they will observe different images 
and consequently recognize the binocular disparity.
Moreover, as the eyes move horizontally,
the observed images change slightly,
realizing so-called motion parallax. 
The binocular disparity and the motion parallax are
two most important visual cues for human 3D perception \cite{dodgson2002analysis}.

In terms of hardware, auto\-stereoscopic displays are easy to implement.
We only need to attach optical elements
in front or at the back of a usual 2D display panel.
Due to the simplicity as well as the cost efficiency,
the auto\-stereoscopic displays have been adopted
in a variety of commercial products
from 3D display walls to mobile devices \cite{benzie2007survey}.

\begin{figure*}[t]
\centering
\includegraphics[width=\textwidth]{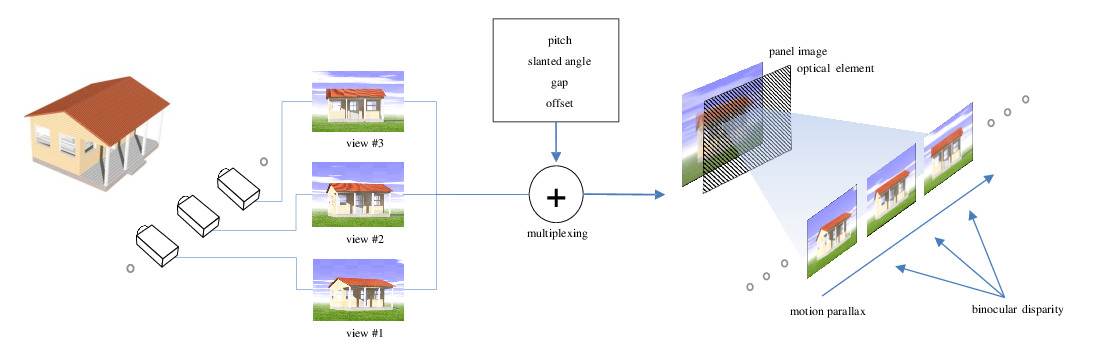}
\caption{Overall procedure of realizing 3D scenes 
in a multi\-view auto\-stereoscopic display.
In this example, we assume that
the input images are from a multiple camera array.}
\label{fig:3D-overall}
\end{figure*}

Despite all the advantages,
several issues still remain \cite{winkler2013stereo}.
Above all, visual quality degradation may happen
by optical misalignment and/or by crosstalk.
If the optical elements are misaligned 
from the desired parameter values (e.g., slanted angle, pitch, etc.), 
the observed images get distorted.
For the auto\-stereoscopic display, 
multiple viewpoint images must be prepared,
by various means \cite{mueller2008view}
(e.g. multiple camera array, depth image based
rendering, image based rendering), 
and be multiplexed together in a panel image,
as shown in Fig.~\ref{fig:3D-overall}. 
Here, the multiplexing must follow
correct view assignment
which depends on the display parameters \cite{ber99}.
In practice, however, the parameters are likely to differ
from the designed values for many reasons: 
fabrication/assembly inaccuracy, thermal effect, 
chronological change, to name a few.
Whatever the reason is, even a small error may lead to
a significant amount of image distortion.

On the other hand,
crosstalk generally refers to a phenomenon whereby
light from a pixel, designated for a specific view point,
is smeared into adjacent ones.
It occurs mainly due to the shape mismatch between the pixels
and the optical elements.
An auto\-stereoscopic display typically slants
the lens or barrier to mitigate moir\'{e} artifacts 
and to divide the resolution reduction by both dimensions.
As a result, rectangular pixels can hardly pass through the slanted
optical element as a whole;
only a fraction can make it, forming a parallelogram
jointly with multiple pieces of other pixels. 
This makes the viewer see not only the intended pixels
but also unintended ones.
Consequently, images may look blurry or look
as if they were multiply-exposed \cite{jarvenpaa2008optical}.
Strictly speaking, given the slanted structure,
the crosstalk is inevitable.
But recent research results \cite{li2011image,zhou2014unified,wang2014improved,lzq+15} show
that one may reduce
the crosstalk ``effects" (i.e., blur, multiple-exposure, etc.)
by appropriately adjusting pixel values 
so that they compensate for the crosstalk.
To do this, one needs a precise model
of the crosstalk which again requires the knowledge of 
the actual display parameters.

To summarize, knowing
the display parameters to a sufficient level of accuracy is prerequisite 
to solving the image distortion and crosstalk problems. 
Therefore, we need to estimate the display parameters at least
once, after installing the optical elements 
to the display, or more frequently
(e.g., on-the-fly) whenever it is necessary.

In this paper, we propose a novel method
for the display parameter estimation.
The proposed method is based on 
the analysis of pattern images observed
from a couple of view points.
We investigate what the observed patterns
tell us about the display parameters
and develop a frequency-domain algorithm
which robustly works in everyday environment.
Unlike prior work, which usually requires
many images to be photographed from all viewing zones
at the optimal viewing distance,
our method only needs two observations
and is largely insensitive to the observation positions.

The remainder of this paper is organized as follows.
We review the related work in the next section.
In Section~\ref{sec:model},
we establish an observation model
for images which are seen through the optical elements
of the auto\-stereoscopic display.
Section~\ref{sec:calib} presents a novel display
calibration method, based on the observation model,
which robustly estimates the display parameters
in frequency domain.
Then, we experimentally validate the proposed method
in Section~\ref{sec:lab}.
Finally, Section~\ref{sec:concl} concludes the paper.

\section{Related Work}
\label{sec:prior-work}

In recent years, plenty of 3D research has focused on
the visual quality improvement.
We first review several work seeking 
to minimize the 3D image distortion
that occurs due to optical misalignment.

Wang {\em et al.} \cite{wang2011cross} propose a method
to correct the sub\-pixel positions of the panel image
when the amount of optical misalignment is known.
The correction is basically equivalent to newly assigning views
to all sub\-pixels according to the actual parameters.
The results demonstrate the quality improvement of the 3D images;
however, the misalignment needs to be somehow measured in advance.
Lee and Ra \cite{lee2006image} measure the misalignment.
They display a periodic color pattern
such that the sub\-pixels of the same view
have the same color,
initially using the designed parameters
for the view assignment.
Then, they observe the pattern 
at the optimal viewing distance.
If the actual parameters differ from the designed ones,
color variations happen.
The authors argue that the misalignment errors are specifically
related to the number of color variations.
They estimate the errors 
based on the counted number of color variations
and re-synthesize the panel image following
the estimated parameters. 
In another work, Lee and Ra \cite{lee2009new} attempt to find
where, among all viewing zones, each sub\-pixel looks brightest
and then to conduct view assignment accordingly.
This approach is supplemented by Zhou {\em et al.}
\cite{zhou2014unified}, who concretely show how to
find the pixel correspondences between the panel and
captured images using structured lights.
The scheme can deal with complicated types of misalignments,
including inhomogeneous ones,
but is not convenient to apply in practice
because it requires too many images
to be photographed from all viewing directions.
Ge {\em et al.} \cite{ge2005camera} propose a computer-vision based method
to estimate the parameters of Varrier\texttrademark,
a huge auto\-stereoscopic display
tiled with thirty-five panels and as many parallax barriers.
Stereo cameras examine left and right images,
mimicking the human eyes.
Then, the parameters are sequentially adjusted
until the cameras find little artifact in both images. 
Hirsch {\em et al.} \cite{hirsch2013construction} show that
the moir\'{e}-magnifier effect 
may be useful for calibrating the angle
between the lens array and the panel.
They establish a mathematical model of the observation 
when a specific pattern is observed through lenses.
We pursue a similar course in this paper.
But Hirsch {\em et al.} do not show
how to automate the calibration
and how to deal with all types of display parameters.
Here our approach departs from theirs.

Another line of research improves the 3D visual quality
by reducing the crosstalk effect.
Li {\em et al.} \cite{li2011image} set up a set of linear equations,
each of which represents the intensity mixture
of light for each site of sub\-pixels.  
They assume that a sub\-pixel, at its intended view point,
is only affected by 
two adjacent sub\-pixels of the same color.
To obtain the equation coefficients,
they measure the crosstalk from the neighbors
using an imaging photometer.
By solving the linear equations,
they find a new intensity value for every sub\-pixel
that effectively cancels out the crosstalk effect.
Similar approaches follow.
Zhou {\em et al.} \cite{zhou2014unified} consider
the same crosstalk model,
but they impose the range constraint
(i.e., between 0 and 255) on the new intensity values.
There might be no solution that exactly
satisfies all the linear equations
as well as the range constraint.
In this reason, Zhou {\em et al.} formulate the crosstalk cancellation
as a constrained least-square problem.
To find the solution,
they use an iterative algorithm.
Wang and Hou \cite{wang2014improved} do pretty
much the same things.
But notably they compute, rather than measure,
the equation coefficients by the visible proportion
of each sub\-pixel.
In more recent work \cite{lzq+15},
Li {\em et al.} still assume the same crosstalk model. 
Instead of imposing the range constraint on the output
(i.e., new intensity value),
they propose a way to estimate the maximum input range
that would keep the output values to be 
within the valid range.
They pre-compress the input dynamic range and simply
solve the linear equations.
Here, for the computation efficiency,
they use inverse filtering, in frequency domain,
which exploits the shift-invariance property of
the crosstalk model.

Note that, in the crosstalk model,
the linear equations actually depend on 
the display parameters.
To compensate for the crosstalk based on the linear equations,
we require them to be sufficiently accurate.
If not, any attempt for crosstalk removal would not only fail
but would also add new artifacts,
which strongly necessitates reliable display parameter estimation.

\section{Observation Model}
\label{sec:model}

\begin{figure}[t]
\centering
\includegraphics[width=.4\textwidth]{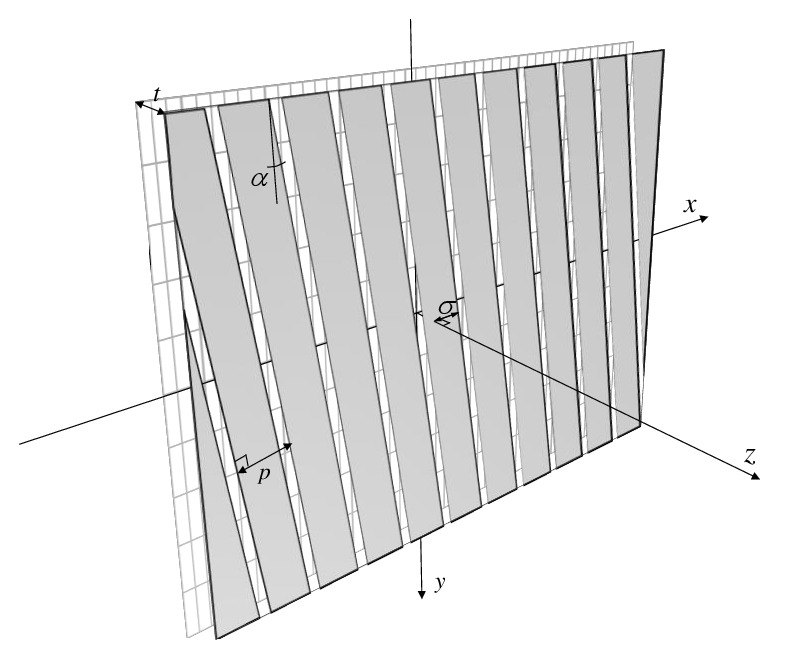}
\caption{3D display with a parallax barrier. 
The architecture is parameterized by the barrier slit pitch $p$;
the slanted angle $\alpha$ of the slits;
the gap $t$ between the panel and the barrier;
and the horizontal offset $\sigma$
by which the slits are located from the center of the panel.}
\label{fig:projective-model}
\end{figure}

In this section, we present a mathematical model 
for the observation when an image on the panel
goes through the optical elements
of the auto\-stereoscopic display.
For clarity, we particularly consider
an auto\-stereoscopic display 
in which parallax barriers exist in front of the panel.\footnote{Our
subsequent results are not restricted to such type of displays.
They are generally applicable to the displays
with rear barriers or with lenticular lenses.}
The display has the following four parameters 
(see Fig.~\ref{fig:projective-model}):
\begin{enumerate}
\item[(1)] Pitch $p$; 
\item[(2)] Slanted angle $\alpha$;
\item[(3)] Gap or Thickness $t$;
\item[(4)] Offset $\sigma$.
\end{enumerate}
We denote the set of the parameters by $\Theta$, 
i.e., $\Theta=(p, \alpha, t, \sigma)$.
For later use, we also define $h$ and $\rho$ as
\begin{align}
h &= \frac{d}{d-t} \, \frac{p}{\cos\alpha}
\label{eq:pitch-relation} \\
\rho&=\frac{d}{d-t} \, \sigma,
\label{eq:offset-relation}
\end{align}
where $d$ is the distance from the panel to the observer's eye.
We will use $\Theta$
to denote the derivative parameters $(h, \alpha, \rho)$,
not only to denote the primary parameters $(p, \alpha, t, \sigma)$,
if it makes no confusion. 

\subsection{Visible Pixels}
\label{sec:visibility}

The parallax barriers enable 
every pixel on the panel to have directionality.
Based on a projective model, we trace back all the rays
that reach the eye position $(u,v)$ at the distance $d$.
Here, for simplicity, we assume a conceptually ideal environment.
The panel has infinite size and infinitely high resolution;
the barrier slits are infinitesimally narrow;
and light does not disperse.

In Fig.~\ref{fig:projective-model},
we can express the barrier slits as a set of lines, i.e., 
\begin{align}
S(\Theta)= \left\{ (x',y'): 
x' = y'\tan\alpha + \sigma + n\frac{p}{\cos\alpha}, 
\enskip \forall n\in {\mathbb Z}\right\}.
\label{eq:slit-function}
\end{align}
The ray that passes both the slit $(x',y')$ and the eye position $(u,v)$
must come from a pixel $(x,y)$ such that
\begin{align}
\left(\begin{matrix}x \\ y\end{matrix}\right)
= \frac{d}{d-t}\left(\begin{matrix}x' \\ y'\end{matrix}\right) 
- \frac{t}{d-t}\left(\begin{matrix}u \\ v\end{matrix}\right).
\label{eq:proj-model}
\end{align}
By arranging Eq.~(\ref{eq:proj-model})
with respect to $(x',y')$ and subsequently by plugging it
into Eq.~(\ref{eq:slit-function}), we obtain
the set of pixels that are visible from the eye at $(u,v)$:
\begin{multline}
P_\gamma(\Theta)=\Bigl\{(x,y): x = y\tan\alpha + \rho  \\
+\Bigl(n+\gamma(u,v;\Theta)\Bigr)h, \enskip \forall n\in {\mathbb Z}\Bigr\},
\label{eq:visible-pixels}
\end{multline}
where 
\begin{align}
\gamma(u,v;\Theta) = 
\frac{t}{pd}\Bigl(v\sin\alpha-u\cos\alpha\Bigr).
\label{eq:eye-position}
\end{align}
The eye position is only one-dimensionally parameterized,
i.e., via $\gamma$.
This is because we assumed a line-type barrier
that only allows one-dimensional parallax.\footnote{The same
is also true for the cylinder-type lenslet array.}
Multiple eye positions have a common set of visible pixels
if they share the same value of $\gamma$.
In this reason, we can use $\gamma$, what we call {\em view},
in place of the eye position $(u,v)$,
whenever the exact 2D coordinate is not necessary.
Note also that $P_\gamma$ is periodic,
in terms of $\gamma$, with the period equal to 1.
A space of the 2D eye positions,
which forms a single period of $\gamma$,
i.e., $\gamma \in [0,1)$,
is often called the {\em primary field of view}.
Beyond it, the set of visible pixels 
simply repeats, extending the effective field of view \cite{ddd+14}.

We have figured out which pixels are visible at the observer's side.
For correct rendering, at the panel's side,
the same set of pixels must be controlled
to display the correspondent image.
This naturally generates the following pixel-view assignment rule:
\begin{align}
R: P_\gamma(\Theta) \mapsto \gamma. 
\label{eq:view-assignment-map}
\end{align}
We can rewrite Eq.~(\ref{eq:visible-pixels}) as
\begin{align}
P_\gamma(\Theta) = \left\{(x,y):
\frac{x-\rho-y\tan\alpha \enskip ({\rm mod}\; h)}{h} = \gamma
\right\},
\label{eq:view-assignment}
\end{align}
by eliminating $n$.
The resulting pixel-view assignment function has exactly the same
form as van Berkel's \cite{ber99}.
Van Berkel actually derived the function 
with lenticular lenses,
while we did with parallax barriers. 
In this sense, our derivation complements his work.

\subsection{Misalignment Effects}
\label{sec:misalignment-effects}

We emphasize that
the display parameters must be known
for the pixel-view assignment.
We denote, by $\Theta_r=(p_r,\alpha_r,t_r,\sigma_r)$ or
$\Theta_r=(h_r,\alpha_r,\rho_r)$, the input parameters
with which the 3D rendering module performs the view assignment.
Let us partition the pixels on the panel by the assigned view.
Then, the set of pixels assigned for view $\gamma'$ becomes equal
to $P_{\gamma'}(\Theta_r)$ (see Eq.~\ref{eq:view-assignment-map}).
The 3D rendering works correctly if,
for every view $\gamma\in[0,1)$,
all the visible pixels $P_\gamma(\Theta)$ are assigned
with the correct view number.
We mathematically express this correct 3D rendering condition as
\begin{align}
P_{\gamma'}(\Theta_r) \cap P_{\gamma}(\Theta) 
=\left\{\begin{array}{ll}
P_{\gamma}(\Theta), & \gamma' = \gamma \\
\varnothing, & \gamma' \neq \gamma
\end{array}\right.
\end{align}
for every $\gamma, \gamma'\in [0,1)$.
One can easily verify that
this condition is simply equivalent to that $\Theta_r=\Theta$.

\begin{figure}[t]
\centering
\includegraphics[width=.35\textwidth]{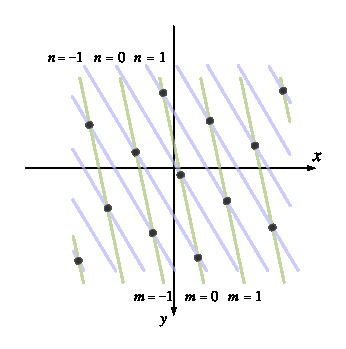}
\caption{2D lattice structure generated by the intersection
of $P_{\gamma'}(\Theta_r)$ (greenish lines),
and $P_{\gamma}(\Theta)$ (blueish lines).
We assume that $\alpha_r\neq \alpha$.
The coordinates of each point are given
by Eqs.~(\ref{eq:2d-lattice-x}),~(\ref{eq:2d-lattice-y}). 
}
\label{fig:2d-lattice}
\end{figure}
What would result if $\Theta_r\neq\Theta$?
We will evaluate $P_{\gamma'}(\Theta_r) \cap P_{\gamma}(\Theta)$
to see which pixels in the observation are actually correct
and which are wrong.
Recall that both $P_{\gamma'}(\Theta_r)$ and $P_{\gamma}(\Theta)$
represent a bunch of lines parallel and equi\-spaced within each set.
We still assume the same idealized environment
that we used in identifying visible pixels: 
panel with infinite resolution, non\-dispersive light, etc.
We will relax the conditions in Section~\ref{sec:calib}. 
If the lines in both sets 
have ``exactly" the same slope (i.e., $\alpha_r=\alpha$),
the intersection can form either another set of lines
or the empty set. 
In this paper, we skip considering this case
because it is unlikely to happen in practice.

If the lines in $P_{\gamma'}(\Theta_r)$ 
have a different slope from those in $P_{\gamma}(\Theta)$
(i.e., $\alpha_r\neq\alpha$),
the intersection produces a set of points:
\begin{align}
x&=\frac{mh_r\tan\alpha-nh\tan\alpha_r}{\tan\alpha-\tan\alpha_r} \nonumber \\
&\hspace*{4em}+\frac{(\rho_r+\gamma' h_r)\tan\alpha-(\rho + \gamma h)\tan\alpha_r }{\tan\alpha-\tan\alpha_r} \label{eq:2d-lattice-x}  \\
y&=\frac{mh_r-nh+\rho_r + \gamma' h_r - \rho - \gamma h}{\tan\alpha-\tan\alpha_r},
\label{eq:2d-lattice-y} 
\end{align}
where $(m, n) \in {\mathbb Z}\times{\mathbb Z}$.
We make the following observations: \\

{\em Lattice encoding.} \enskip The points form a lattice pattern 
in 2D plane (see Fig.~\ref{fig:2d-lattice}).
The lattice pattern encodes the actual parameters
$\Theta=(h,\alpha,\rho)$ in a certain way. \\[-.5em]

{\em View-invariant structure.} \enskip The lattice structure is invariant
(only up to a shift) to both view parameters $\gamma$ and $\gamma'$.\\[-.5em]

{\em Mixed views.} \enskip 
With $\gamma$ fixed and $\gamma'$ varying (from 0 to 1),
the lattice pattern makes a continuous shift
along the lines of $P_\gamma(\Theta)$.
This implies that rainbow-like view blending will appear in the observation.
This phenomenon quite much resembles crosstalk.
But remember that it happens due to
the wrong assignment of views by the rendering module
rather than due to light leakage.\footnote{Certain studies
(e.g., \cite{zhou2014unified}) define ``crosstalk" in wide terms,
while classifying it into {\em intrinsic} crosstalk
and {\em extrinsic} crosstalk.
What we consider here corresponds to the extrinsic one.}

\section{Calibration}
\label{sec:calib}

Supposing that we do not know the actual display parameters,
how can we estimate them with a high level of accuracy?
In this section, we provide a method,
based on the observation model that we considered in the previous section. 

We seek to obtain all physical parameters,
i.e., $p,\alpha,t,\sigma$, in our calibration.
However, a single observation is not sufficient
for unambiguous identification.
We may easily show this by example.
Consider the following two cases:
(1) $p=0.9975\times\sqrt{10}$, $\alpha=\arctan(1/3)$, $t=25$, $\sigma=0$,
(2) $p=0.9950\times\sqrt{10}$  $\alpha=\arctan(1/3)$, $t=50$, $\sigma=0$,
while the observer is commonly at $(u,v)=(0,0)$,
distant from the panel by $d=10,000$.  
If we evaluate $P_\gamma(\Theta)$ in Eq.~(\ref{eq:visible-pixels}),
it is exactly the same for both cases.
Given the same image,
the observer sees exactly the same thing for both cases,
not being able to distinguish one case from the other.
In this study, 
we make a couple of observations
to get rid of the ambiguity.

For the automated calibration,
we use a camera as the measurement device.
In this context, hereafter,
the position $(u,v)$, view $\gamma$, and the distance $d$ 
denote those of the camera, not of a human eye.
We assume that the camera coordinate $(u,v,d)$ can be known
from state-of-the-art camera pose estimation techniques
(e.g., \cite{malis2007deeper}).
Section~\ref{sec:camera-calib} explains the procedure.

Although many recent studies have been proposed
for the display calibration,
they generally require that a number of observations
should be made (see Section~\ref{sec:prior-work}).
A few exceptions exist. 
The work by Hirsch {\em et al.} \cite{hirsch2013construction} 
and by Lee and Ra \cite{lee2006image}
needs rather a small number of observations.
However, the aim of \cite{hirsch2013construction} is basically
to help a ``human" observer 
to calibrate the angle ($\alpha$ in our notation) in micro levels,
so it does not fit our purpose (i.e., automated calibration
dealing with all types of display parameters). 
In contrast, \cite{lee2006image}
can be made automated and be made applicable, without difficulty,
for finding all display parameters.
But its drawback lies in the accuracy.
Their model, based on counting color change,
suffers from a sort of quantization effect.
The counting-based measurements are not easy to
obtain to a small fractional precision.
Besides, their modeling holds only if
the actual display parameters are sufficiently close
to the designed ones (or the initial ones).
Otherwise, aliasing may happen and every color change 
may not be observed, violating their assumption.

\subsection{Pattern Images}
\label{sec:pattern-images}

In our study, we seek to directly observe the lattice encoding pattern
(as in Fig.~\ref{fig:2d-lattice})
without complication.
For this purpose, we prepare the all-white image for a single view $\gamma'=0$
and all-black images for the other views, and
interweave them according to our rendering parameters.
The design parameters may be available,
and we may want to use them for rendering.
However, in our perspective,
they are merely one of many values
that mismatch the actual parameters.
Instead of using those values, we choose rendering parameters
such that it can make our analysis as simple as possible.
In this study, we use 
\begin{align}
h_r=\beta, \enskip \alpha_r=0, \enskip \rho_r=\epsilon
\label{eq:rendering-choice}
\end{align}
with $\beta$ being a multiple of 3.
With the choice, 
the panel image is filled with vertical stripes
of the same color, spaced by $\beta$
and offset by $\epsilon$.
Then, the lattice pattern 
is reduced to (see Eqs.~\ref{eq:2d-lattice-x}~and~\ref{eq:2d-lattice-y})
\begin{multline}
L_{\gamma}(x,y)=\sum_{m=-\infty}^\infty\sum_{n=-\infty}^\infty
\Biggl\{\delta\left(x-m\beta-\epsilon\right) \\
\cdot\delta\left(y-m\frac{\beta}{\tan\alpha}+n\frac{h}{\tan\alpha}-\tau\right)
\Biggr\},
\label{eq:L-gamma}
\end{multline}
where 
\begin{align}
\tau=\frac{\epsilon-\rho-\gamma h}{\tan\alpha}.
\label{eq:tau-rho-gamma}
\end{align}
In (\ref{eq:L-gamma}), $\delta$ denotes Dirac's delta function.

\subsection{Display Parameter Estimation}
\label{sec:param-estim}

We have assumed several ideal conditions
in deriving the lattice pattern $L_\gamma(x,y)$.
Now, let us relax them one by one.\\

(i) {\em The panel has a finite size.} \enskip We denote 
the finite window by $B(x,y)$,
which would typically be a rectangular function 
without special treatment. \\[-.5em]

\begin{figure}[t]
\centering
\subfigure[]{
\includegraphics[width=.225\textwidth]{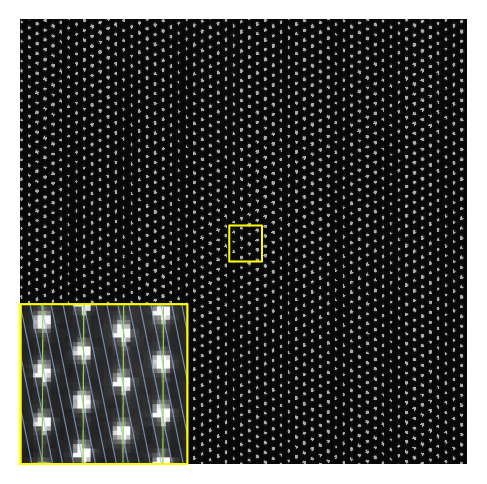}
\label{fig:2d-lattice-in-reality-spatial}
}
\subfigure[]{
\includegraphics[width=.225\textwidth]{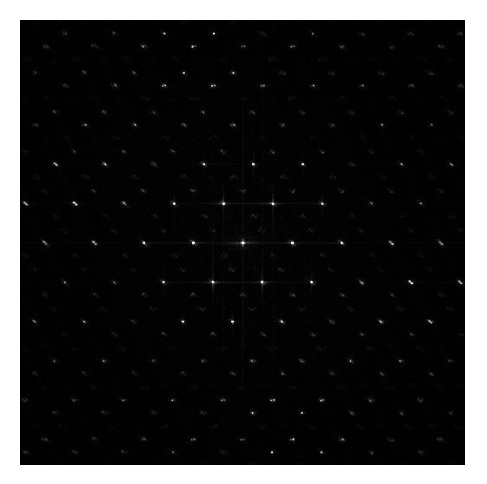}
\label{fig:2d-lattice-in-reality-frequency}
}
\caption{2D lattice pattern found in an actual observation.
\subref{fig:2d-lattice-in-reality-spatial} Spatial domain, 
\subref{fig:2d-lattice-in-reality-frequency} Frequency domain.
The boxed region is shown in magnification,
with $P_{\gamma'}(\Theta_r)$ and $P_{\gamma}(\Theta)$ overlaid together
(in greenish and blueish color, respectively), for better viewing.
Note that a 2D lattice in spatial domain is mapped to
another 2D lattice, called {\em reciprocal lattice} \cite{kit05},
in frequency domain.
}
\label{fig:2d-lattice-in-reality}
\end{figure}

(ii) {\em Barrier slits have a certain width and light may disperse.}
\enskip 
The points on the panel spread (see Fig.~\ref{fig:2d-lattice-in-reality}).  
We denote the point spread function by $H(x,y)$.
The shape actually depends on the slit width,
slanted angle $\alpha$, distance $d$, etc.,
so $H(x,y)$ is difficult to specify accurately.
However, it commonly delivers
some low-pass filter (LPF) characteristics. \\[-.5em]

(iii) {\em The observation tends to be contaminated by noise.} \enskip 
We assume, for simplicity, that the noise $N(x,y)$
is independent and identically distributed (i.i.d.) over the pixels. \\[-.5em]

(iv) {\em The panel has a finite resolution.} \enskip 
This makes the actual observation be a sampled version
of the ideal observation. We postpone considering this non\-ideality
to a little later part. \\

Taking the first three non\-idealities into accounts,
we may express the actually observed image $I_\gamma$ as
\begin{align}
I_\gamma(x,y)=B(x,y)\Bigl(H(x,y) \star L_\gamma(x,y)\Bigr)+N(x,y),
\label{eq:observation}
\end{align}
where $\star$ denotes the convolution operator.
If we define $L_o(x,y)$ as
\begin{multline}
L_o(x,y)=\sum_{m=-\infty}^\infty\sum_{n=-\infty}^\infty
\Biggl\{
\delta\left(x-m\beta\right) \\
\cdot\delta\left(y-m\frac{\beta}{\tan\alpha}+n\frac{h}{\tan\alpha}\right)
\Biggr\},
\end{multline}
$L_\gamma(x,y)$ is merely a shift of $L_o(x,y)$,
horizontally by $\epsilon$ and vertically by $\tau$ (This is
due to the view invariance property of the lattice structure;
see Section~\ref{sec:misalignment-effects}).
Therefore, we can rewrite Eq.~(\ref{eq:observation}) as
\begin{align}
I_\gamma(x,y)=B(x,y)\Bigl(H(x,y) \star L_o(x-\epsilon,y-\tau)\Bigr)+N(x,y),
\label{eq:observation-2}
\end{align}
where $\tau$ is a function of $\gamma$ as given by Eq.~(\ref{eq:tau-rho-gamma}). 

We further eliminate the shift variation by moving
our analysis to the frequency domain.
As is well known, a shift leads to the phase modulation 
in frequency domain and does not affect the signal magnitude.
In the frequency domain,
Eq.~(\ref{eq:observation-2}) is transformed into
\begin{multline}
\hat{I}_\gamma(f_x,f_y)=e^{-2\pi (f_x\epsilon+f_y\tau)}\hat{B}(f_x,f_y) \\ \star\left(\hat{H}(f_x,f_y)
\hat{L}_o(f_x,f_y)\right) 
+\hat{N}(f_x,f_y),
\label{eq:observed}
\end{multline}
where we used the multiplication-convolution duality.

Generally, a 2D lattice in spatial domain is mapped to
another 2D lattice in frequency domain (see \cite{kit05,agn04} for reference). 
In our case, $\hat{L}_o(f_x,f_y)$ is given by
\begin{multline}
\hat{L}_o(f_x,f_y) = C
\sum_{m=-\infty}^{\infty}\sum_{n=-\infty}^{\infty}
\Biggl\{\delta\left(f_x-m\frac{1}{\beta}+n\frac{1}{h}\right) \\
\cdot\delta\left(f_y-n\frac{\tan\alpha}{h}\right)\Biggr\},
\label{eq:hat-Lo}
\end{multline}
where $C$ denotes a constant scale factor.
If we insert Eq.~(\ref{eq:hat-Lo}) into Eq.~(\ref{eq:observed})
and make a bit of arrangement, we obtain
\begin{multline}
\hat{I}_\gamma(f_x,f_y)=\sum_{m,n} 
w_{mn} \hat{B}\left(f_x-\frac{m}{\beta}+\frac{n}{h},
f_y-\frac{n\tan\alpha}{h}\right) \\
+\hat{N}(f_x,f_y),
\label{eq:observed-final}
\end{multline}
where 
\begin{align}
w_{mn}=Ce^{-2\pi i m\frac{1}{\beta}\epsilon}
e^{-2\pi i n(\frac{\tan\alpha}{h} \tau-\frac{1}{h}\epsilon)}
\hat{H}\left(\frac{m}{\beta}-\frac{n}{h},\frac{n\tan\alpha}{h}\right).
\label{eq:mixture-coefficient}
\end{align}
We interpret Eq.~(\ref{eq:observed-final}) as follows:
The signal part forms a mixture of functions $\hat{B}$.
The functions would be sincs if $B(x,y)$ were a rectangular window.
We use a simple trick to make $B(x,y)$ be Gaussian.
We multiply Gaussian weights to the pixel values.
Then, in our case, the signal part of $\hat{I}_\gamma$ 
becomes a mixture of Gaussians,
rather than a mixture of sincs,
because a Gaussian remains as a Gaussian
(with a different bandwidth) in frequency domain.
The mixture weights $w_{mn}$ are proportional to $\hat{H}$,
which has the LPF characteristics as aforementioned.
This means that the amplitude of a Gaussian is high around the center
of the frequency axes and diminishes away from the center. 

Finally, we attempt to consider the remaining non\-ideality
that the panel has a finite resolution.
In the frequency domain, this poses two implications.
First, the signal replicates due to the sampling effect.
The replica might be able to corrupt the original signal,
behaving like noise.
But recall that the Gaussians far from the origin
have attenuated weights. When they are replicated
at the multiples of the sampling frequency 
and then stretched to around the origin,
the magnitudes become negligible.
Therefore, even with the sampling effects,
Eq.~(\ref{eq:observed}) holds around the origin quite well.
Second, the resolution of the observation may be so low 
that we cannot directly extract the parameters
from the observation,
to a sufficient level of accuracy.
We will soon show how to deal with this issue 
in our calibration. \\

\begin{figure}
\centering
\subfigure[]{
\includegraphics[width=.2\textwidth]{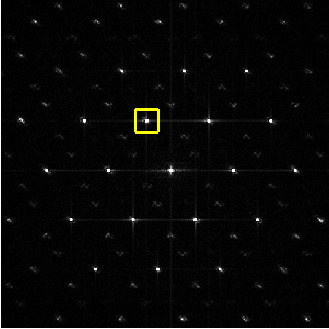}
\label{fig:peak-localization-full}
}
\\
\subfigure[]{
\includegraphics[width=.2\textwidth]{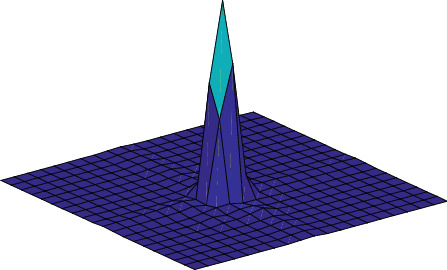}
\label{fig:peak-localization-magnified}
}
\quad
\subfigure[]{
\includegraphics[width=.2\textwidth]{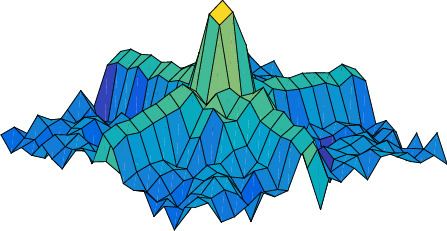}
\label{fig:peak-localization-magnified-frequency}
}
\caption{Accurate peak localization. 
\subref{fig:peak-localization-full} Initial ballpark peak point and the neighborhood. 
\subref{fig:peak-localization-magnified} Magnified plot.
The shape around the peak is almost a (coarsely sampled) Gaussian.
\subref{fig:peak-localization-magnified-frequency} Magnified plot in log domain.
The shape around the peak is almost a (coarsely sampled) paraboloid.
The peak point is estimated by the paraboloid fitting
in the log domain.
}
\label{fig:peak-localization}
\end{figure}

{\bf Horizontal pitch $h$, Slanted angle $\alpha$: }
Given $\hat{I}_\gamma(f_x,f_y)$, we detect
a Gaussian peak around the origin.
To accurately localize the peak, we employ paraboloid fitting
around the initial ballpark.
The paraboloid fitting can be performed
quite efficiently because the closed-form solution is available.
A similar approach was proposed by Cho {\em et al.}
\cite{cho2013modeling} in calibrating light-field cameras.
But our scheme includes a notably distinct feature.
Recall that our signal around its peak is almost a Gaussian
because we cooked it by means of the window function $B(x,y)$.
In log domain, it ``indeed" becomes a paraboloid 
(see Fig.~\ref{fig:peak-localization}), so  
the paraboloid fitting in the log domain
should work very well.

With no noise involved, this provides us 
\begin{align}
f_x^*=\frac{m}{\beta}-\frac{n}{h}, \quad
f_y^*=\frac{n\tan\alpha}{h}
\label{eq:measurements}
\end{align}
for some $m, n \in {\mathbb Z}$.
In fact, noise is well suppressed in frequency domain.
We will argue that the signal-to-noise ratio (SNR)
around a Gaussian peak is quite high in frequency domain.
To see this, let us consider the signal power and the noise power
around peaks.
First, due to the unitary property of the Fourier transform,
the {\em total} signal power is preserved 
(this is known as Parseval's theorem \cite{own96}).
But the number of lattice points is usually far fewer 
in frequency domain than in spatial domain
(e.g., see Fig.~\ref{fig:2d-lattice-in-reality}),
and it implies that the signal power is concentrated
on the points (or peaks).
On the other hand, noise power is evenly spread
over all the frequencies
because the noise is assumed to be i.i.d.\ over the pixels
and the Fourier transform is unitary.
Therefore, the SNR becomes quite high around a peak
in frequency domain (while quite low around a valley).

From Eq.~(\ref{eq:measurements}), we obtain
\begin{align}
h = \frac{n}{m/\beta-f_x^*}, \quad
\alpha=\arctan\frac{f_y^*}{m/\beta-f_x^*}. 
\label{eq:h-alpha}
\end{align}
If $m$ and $n$ are somehow known, 
Eq.~(\ref{eq:h-alpha}) directly gives us the two parameters $h$ and $\alpha$.
Otherwise, $h$ and $\alpha$ remain in uncertainty,
but they are still restricted to be in a discrete set of candidates.
To break the ambiguity, we actually multiplex two patterns
in a panel image (with different values of $\beta$ and $\epsilon$).  
To be specific, we use $\beta=15$, $\epsilon=0$
for one pattern and $\beta=24$, $\epsilon=1$ 
for the other.
Multiplexed together, these patterns generate
the panel image that consists of two sets of stripes, 
each differently colored and differently spaced.
In the observation, the two stripe patterns can easily be separated.
Then, for each pattern $j=1, 2$, we construct the candidate set
${\cal C}_j(h,\alpha)$ using Eq.~(\ref{eq:h-alpha})
for $m, n \in {\mathbb Z}$.
Then, we intersect the candidate sets ${\cal C}_1$, ${\cal C}_2$
to find $h$ and $\alpha$
that can explain both patterns at the same time.
In practice, however, noisy factors or numerical imprecisions are 
likely to make the ``exact" intersection be empty.
So, we actually compute $h$ and $\alpha$ such that
they are close to ${\cal C}_1$ and ${\cal C}_2$ at the same time:
\begin{align}
(h,\alpha)=\argmin_{(h',\alpha')}\min_{(h,\alpha)_j\in{\cal C}_j\atop j=1,2}
\sum_{j=1}^2\,\Bigl\|(h',\alpha')-(h,\alpha)_j\Bigr\|^2. 
\end{align}
\\

{\bf Pitch $p$, Gap or Thickness $t$: }
To further identify the physical parameters $p$ and $t$,
we make a secondary observation at a different distance
from the first one.
Let $d_k$ denote the distance of the camera
from the panel for the $k$th observation ($k=1,2$).
We meant that $d_1 \neq d_2$.
We assume that the distances are known
by a camera estimation technique
(see Section~\ref{sec:camera-calib})
and that
we have already estimated the slanted angle $\alpha$
and the horizontal pitch $h_k$ for each observation $k$.

For each observation,
the horizontal pitch is related to the physical pitch $p$ and gap $t$
by (see Eq.~\ref{eq:pitch-relation})
\begin{align}
h_k=\frac{d_k}{d_k-t}\cdot\frac{p}{\cos\alpha}, \quad k=1,2.
\label{eq:pd-equations}
\end{align}
It is straightforward to solve Eq.~(\ref{eq:pd-equations}).
The solution is given by
\begin{align}
p &= \frac{h_1h_2(d_2-d_1)\cos\alpha}{d_2h_1-d_1h_2}, \label{eq:p-estimate} \\
t &= \frac{d_1d_2(h_1-h_2)}{d_2h_1-d_1h_2}. \label{eq:t-estimate} \\ 
\nonumber
\end{align}

{\bf Offset $\rho$, $\sigma$: }
The offset parameters only affect how much the observed image
$I_\gamma(x,y)$ is shifted from the lattice pattern $L_o(x,y)$.
Given all other display parameters 
as well as the camera pose,
the amount of the shift can be robustly estimated by
finding $\zeta$ that best matches $I_{\gamma}(x+\epsilon,y+\zeta)$
to $L_o(x,y)$.
In frequency domain, the operation is implemented by
\begin{multline}
\tau = \argmax_{\zeta} \; \Biggl|\sum_{m,n} \hat{I}_{\gamma}\left(\frac{m}{\beta}-\frac{n}{h},\frac{n\tan\alpha}{h}\right) 
\\
\cdot e^{2\pi i (m\frac{1}{\beta}\epsilon+ n(\frac{\tan\alpha}{h} \zeta-\frac{1}{h}\epsilon))}\Biggr|.
\label{eq:corr}
\end{multline}
After computing $\tau$, we can estimate the parameter $\rho$ as
(see Eq.~\ref{eq:tau-rho-gamma})
\begin{align}
\rho=\epsilon-\tau\tan\alpha-\gamma h.
\label{eq:rho-estimate}
\end{align}
where $\gamma$ is evaluated in advance,
by Eq.~(\ref{eq:eye-position}),
on the basis of the known parameters.
Then, we can subsequently find the offset $\sigma$,
by Eq.~(\ref{eq:offset-relation}),
\begin{align}
\sigma=\frac{d-t}{d}\rho.
\label{eq:sigma-estimate}
\end{align}

\subsection{Camera Pose Estimation}
\label{sec:camera-calib}

We assume that the intrinsic camera parameters
(e.g., focal length, principal point, radial distortion coefficients) are known.
Otherwise, camera calibration (e.g., see \cite{zhang2000flexible}) can be conducted in advance to the 3D display calibration.

To estimate camera poses,
we exploit a state-of-the-art method in that field,
called {\em homography decomposition} \cite{malis2007deeper}.
We detect four corners of the display in the captured image
and find homography between the world coordinates and image coordinates.
The homography can be decomposed into
the intrinsic parameter matrix and the camera pose matrix.
As assumed, we know the intrinsic parameter matrix,
so we can compute the camera pose matrix
by multiplying its inverse to the homography.
The camera pose matrix tells us the camera rotation
and translation relative to the origin (i.e., panel center)
of the world coordinate.
The translation vector exactly corresponds to
the camera position $(u,v,d)$.

If the camera is rotated,
the shape of the display may appear geometrically distorted
in the captured image.
For our display calibration, we need it to be rectified. 
This could be done with a usual image rectification technique
using the computed rotation matrix.
In this study, we use an equivalent yet simpler scheme
that warps the four corners to the vertices of 
a rectangle of the panel resolution.

\section{Experiments}
\label{sec:lab}

We conduct two types of experiments to evaluate the
performance of the proposed calibration scheme.
In the first set of experiments,
we generate a synthetic dataset 
from POV-Ray simulation \cite{pov-ray},
for diverse ``virtual" 3D displays.
The ground-truth display parameters are available
for the synthetic dataset,
so we can quantitatively evaluate the estimation error.
We also run the experiments in noisy settings
and see how robust the proposed scheme is.
In the second set of experiments,
we apply the proposed scheme to real-life auto\-stereoscopic displays.
For those displays, the actual display parameters are generally unknown,
which makes it difficult to directly compute the estimation error.
In this case, we evaluate the performance
in the visual aspect of the observed images
when rendering is done using the estimated display parameters.
We assess the visual quality in subjective terms as well as
two objective measures (i.e., peak-signal-to-noise ratio,
structural similarity index).\footnote{Some test images are
from public domain (Big Buck Bunny \copyright Blender Foundation,
UCSD/MERL Light Field Repository \cite{ucsd-merl},
The Stanford Light Field Archive \cite{stanford-lf},
MIT Synthetic Light Field Archive \cite{mit-lf}),
and others are graphically generated by ourselves.}

\begin{table}[t]
\centering
\caption[]{3D Displays Implemented in Our Simulation Environment \\[.2em]
{\scriptsize Units in Parameter Spec.: Milli-meters (Length), Degrees (Angle)}
}
\vspace*{-1em}
\label{tab:synthetic-dataset}
\begin{tabular}[t]{ccccccc}
\hline 
\multirow{2}{*}{Display} & Size &
Optical & \multicolumn{4}{c}{Designed parameters} \\ \cline{4-7}
& (Resolution) & elements & $p_o$ & $\alpha_o$ & $t_o$ & $\sigma_o$ \\ \hline\hline
\multirow{2}{*}{FHD55B} & 55\,in & \multirow{2}{*}{Barrier} 
& \multirow{2}{*}{1} & \multirow{2}{*}{18} & \multirow{2}{*}{4} 
& \multirow{2}{*}{0.5} \\  
& (1\,920$\times$1\,080) & & & & &  \\ \hline
\multirow{2}{*}{UHD32B} & 32\,in & \multirow{2}{*}{Barrier} 
& \multirow{2}{*}{0.5} & \multirow{2}{*}{10} & \multirow{2}{*}{2}
& \multirow{2}{*}{0.2} \\ 
& (3\,840$\times$2\,160) & & & & &  \\ \hline
\multirow{2}{*}{WQXGA10L} & 10\,in & \multirow{2}{*}{Lenticular} 
& \multirow{2}{*}{0.1} & \multirow{2}{*}{12} & \multirow{2}{*}{1} 
& \multirow{2}{*}{0.05} \\
& (2\,560$\times$1\,600) & & & & &  \\ \hline 
\end{tabular} 
\end{table}

\subsection{Synthetic Dataset Calibration}
\label{sec:synthetic-lab}

For the synthetic dataset,
we build a simulation environment based on a ray tracing tool called POV-Ray \cite{pov-ray}.
We consider three types of 3D displays
as listed in Table~\ref{tab:synthetic-dataset}
where the designed parameters $\Theta_o=(p_o,\alpha_o,t_o,\sigma_o)$
are shown together.
For misalignment,
we add perturbations to the designed parameters.
We assume that the perturbations are 
uniformly distributed over the following range: 
\begin{multline}
\frac{|p-p_o|}{p_o}<0.01, \enskip 
\frac{|\alpha-\alpha_o|}{\alpha_o}<0.01, \\
\enskip \frac{|t-t_o|}{t_o}<0.01, \enskip 
\frac{|\sigma-\sigma_o|}{p_o}<0.01.
\end{multline}
For each type of display,
we implement 10 instances,
each with independent random perturbations.

We capture two images for each display instance
at different distances.
In the simulation, we place the camera
at some random positions, deviated up to $\pm 50$\,mm
in all directions,
rotated by a small random angle $\varphi$ ($|\varphi|<$1\,$^\circ$)
along a random axis.
The captured images have an ultra-high-definition (UHD) resolution,
i.e., 3\,840$\times$2\,160.
Given two captured images,
we first conduct rectification 
as well as the camera pose estimation
(see Section~\ref{sec:camera-calib}).
The rectified images are moved
to frequency domain, using the discrete Fourier transform,
in which the display parameters are estimated
sequentially through several steps
(see Section~\ref{sec:param-estim}). 
Table~\ref{tab:synthetic-lab} presents the estimation results
and makes comparisons with those 
obtained with the calibration based on prior work \cite{lee2006image}.
We have implemented \cite{lee2006image} by ourselves.
The original scheme, based on counting color changes, can be used
to estimate $h$ and $\alpha$,
but it does not actually give a way to estimate $p$ and $t$. 
In our implementation, we compute $p$ and $t$
just as in the proposed method (following Eqs.~\ref{eq:p-estimate},
~\ref{eq:t-estimate}), on top of their estimate of $h$ and $\alpha$.
For the offset parameter $\sigma$,
we follow their strategy of taking additional pictures
after correcting $p$, $t$, $\alpha$.
Specifically, we only turn on the center view pixels
and photograph the observed images from all viewing zones.
The offset parameter can be easily computed by spotting
which of the pictures is the brightest.
In the table, we observe that the proposed scheme works quite well,
showing several times higher accuracy 
than prior work.
\begin{table}[t]
\centering
\caption[]{Parameter Estimation Errors \\[.2em]
{\scriptsize The Shown Numbers are The Mean and Standard Deviation \\ 
(in Parenthesis) of The Absolute Errors \\ 
Over 10 Instances Per Each Type of Display.}
}
\label{tab:synthetic-lab}
Calibration based on prior work \cite{lee2006image} \\[-0.5em]
\begin{tabular}[t]{ccccc}
\hline 
\multirow{2}{*}{Display} &  
\multicolumn{4}{c}{Estimation errors (length: mm, angle: $^\circ$)} \\ \cline{2-5}
&  $|\Delta p|$ & $|\Delta\alpha|$ & $|\Delta t|$ & $|\Delta\sigma|$ \\ \hline\hline
\multirow{2}{*}{FHD55B} & 0.0099 & 4.18$\times$10$^{-4}$ & 0.1639 & 0.0460 \\  
& (0.0254) & (5.99$\times$10$^{-4}$) & (0.2399) & (0.0154) \\ \hline
\multirow{2}{*}{UHD32B} & 0.0356 & 1.94$\times$10$^{-4}$ & 0.1455 & 0.0173 \\  
& (0.0818) & (3.58$\times$10$^{-4}$) & (0.2680) & (0.0043) \\ \hline
\multirow{2}{*}{WQXGA10L} & 0.0042 & 1.02$\times$10$^{-5}$ & 0.1980 & 0.0048 \\  
& (0.0077) & (7.76$\times$10$^{-6}$) & (0.1340) & (0.0063) \\ \hline
\end{tabular} 
\\[2em]
Calibration based on the proposed method \\[-0.5em]
\begin{tabular}[t]{ccccc}
\hline 
\multirow{2}{*}{Display} &  
\multicolumn{4}{c}{Estimation errors (length: mm, angle: $^\circ$)} \\ \cline{2-5}
&  $|\Delta p|$ & $|\Delta\alpha|$ & $|\Delta t|$ & $|\Delta\sigma|$ \\ \hline\hline
\multirow{2}{*}{FHD55B} & 0.0027 & 4.21$\times$10$^{-5}$ & 0.0296 & 0.0281 \\  
& (0.0011) & (3.96$\times$10$^{-5}$) & (0.0250) & (0.0175) \\ \hline
\multirow{2}{*}{UHD32B} & 0.0017 & 2.98$\times$10$^{-5}$ & 0.0304 & 0.0069 \\  
& (0.0005) & (1.77$\times$10$^{-5}$) & (0.0190) & (0.0040) \\ \hline
\multirow{2}{*}{WQXGA10L} & 0.0061 & 4.80$\times$10$^{-6}$ & 0.1643 & 0.0069 \\  
& (0.0011) & (4.17$\times$10$^{-6}$) & (0.0530) & (0.0019) \\ \hline
\end{tabular} 
\end{table}

\begin{figure*}[t]
\centering
\includegraphics[width=.9\textwidth]{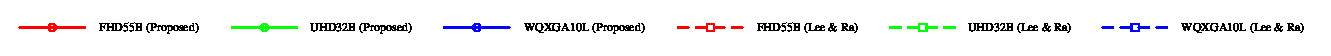}
\\[-.5em]
\subfigure[]{
\label{fig:noisy-lab-pitch}
\includegraphics[width=.23\textwidth]{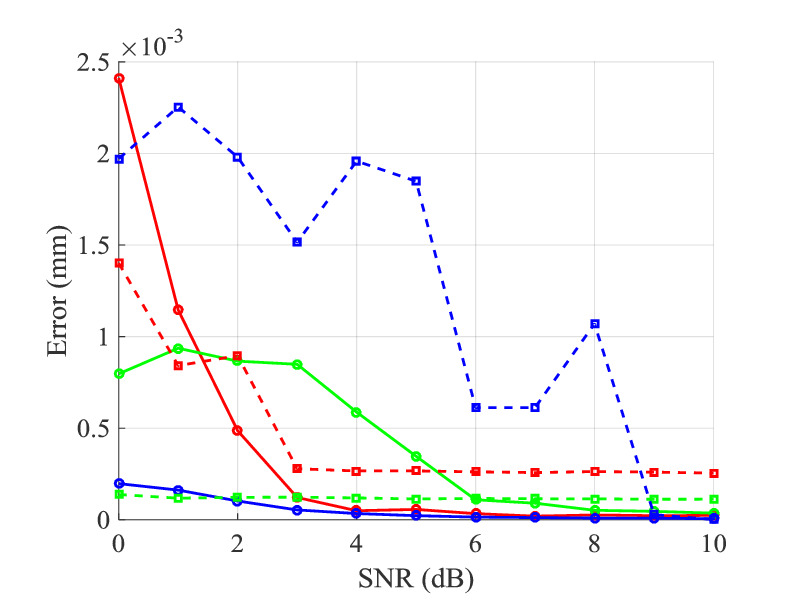}
}
\subfigure[]{
\includegraphics[width=.23\textwidth]{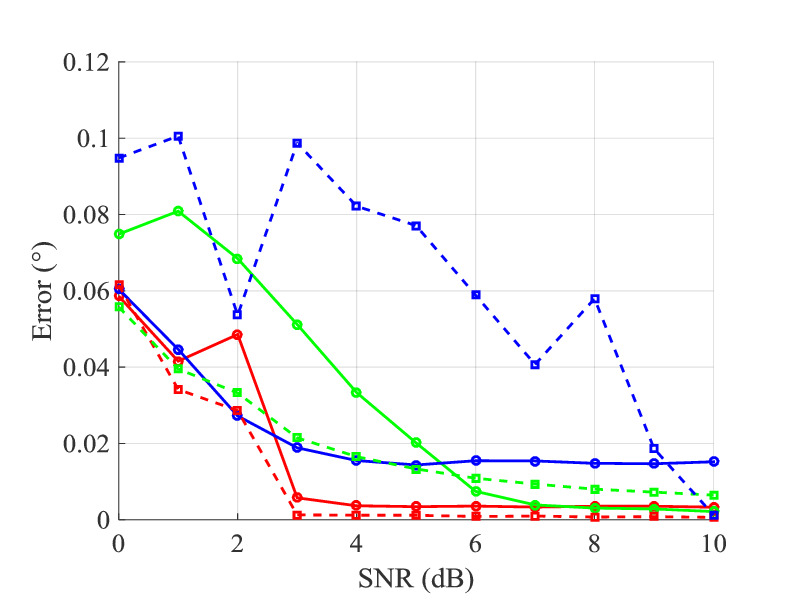}
\label{fig:noisy-lab-angle}
}
\subfigure[]{
\includegraphics[width=.23\textwidth]{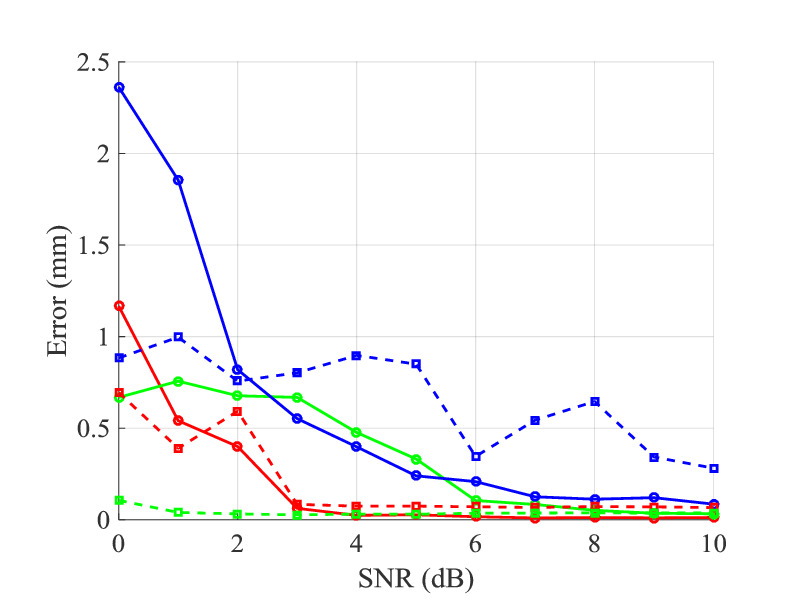}
\label{fig:noisy-lab-gap}
}
\subfigure[]{
\includegraphics[width=.23\textwidth]{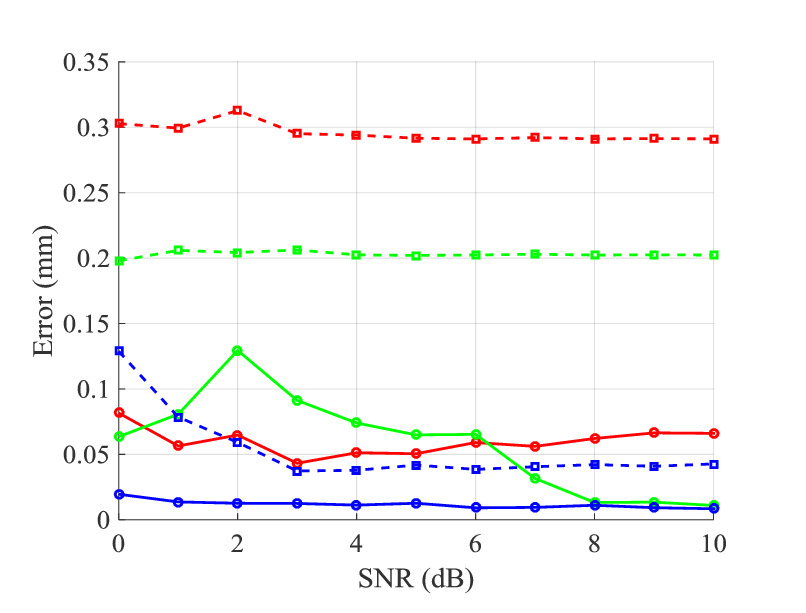}
\label{fig:noisy-lab-offset}
}
\caption{Mean absolute errors for display parameter estimation
in noisy settings.
\subref{fig:noisy-lab-pitch} Pitch $p$, 
\subref{fig:noisy-lab-angle} Slanted angle $\alpha$, 
\subref{fig:noisy-lab-gap} Gap or Thickness $t$,
\subref{fig:noisy-lab-offset} Offset $\sigma$.
For each sub-figure, we compare the performance
between the calibration based on prior work \cite{lee2006image}
and the proposed calibration method.
We find that the errors increase with the noise but not quite rapidly.
}
\label{fig:noisy-lab}
\end{figure*}

We run similar experiments in noisy settings.
We add Poisson noise to simulate
photon shot noise in usual photographic images \cite{st91}.
The results are shown in Fig.~\ref{fig:noisy-lab}.
As expected, the mean absolute errors increase
with the level of noise, but the rate is not high.
The errors are maintained quite low
even with a significant amount of noise.
The offset parameter $\sigma$ turns out 
to be most vulnerable to the noise.
This is probably due to error accumulation;
the estimate of the offset parameter
depends on those of the other parameters
in our calibration.
We plan to handle this issue, error accumulation, in our future work.

\subsection{Real-Life Display Calibration}
\label{sec:real-lab}

We apply the proposed method 
on real-life auto\-stereo\-scopic displays as well.

\begin{figure*}[t!]
\centering
{\small
\includegraphics[width=.1\textwidth]{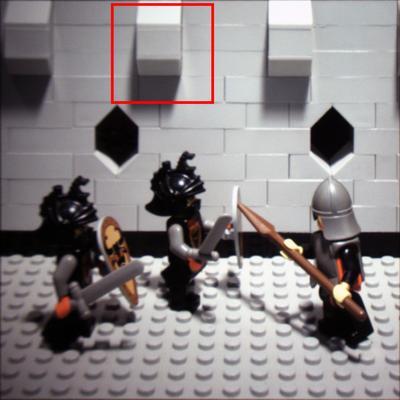}
\includegraphics[width=.1\textwidth]{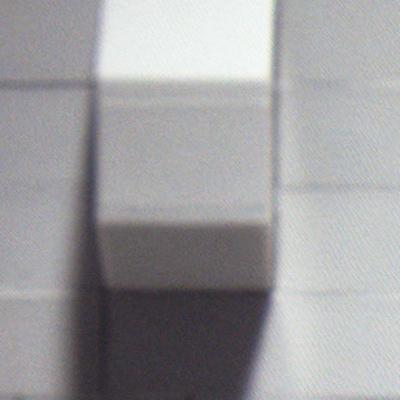}
\quad
\includegraphics[width=.1\textwidth]{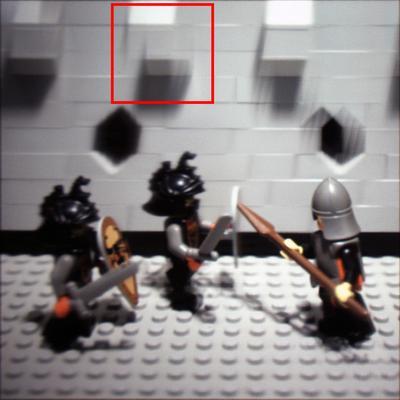}
\includegraphics[width=.1\textwidth]{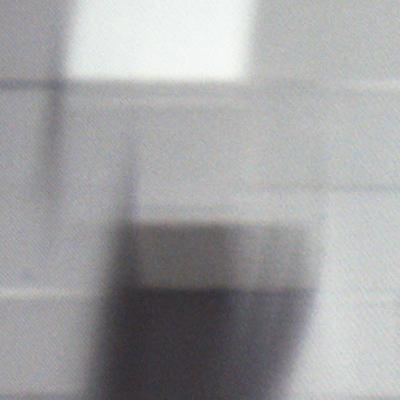}
\quad
\includegraphics[width=.1\textwidth]{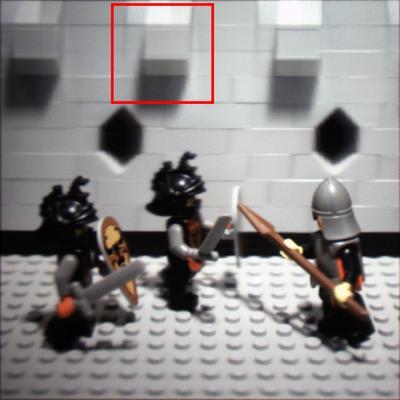}
\includegraphics[width=.1\textwidth]{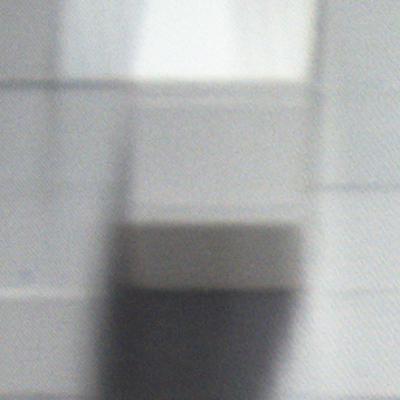}
\quad
\includegraphics[width=.1\textwidth]{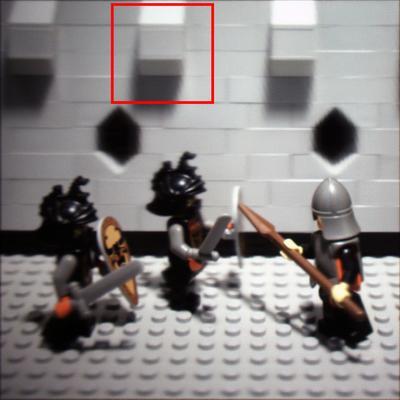}
\includegraphics[width=.1\textwidth]{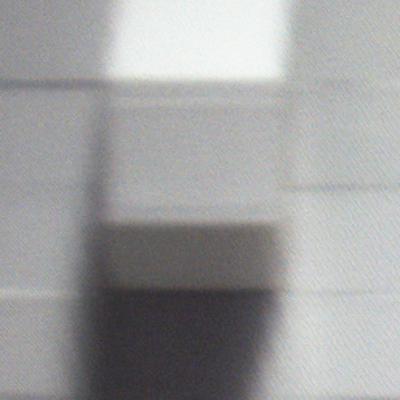}
\\
\hspace*{17.8em}
23.51\,dB (0.73) \hspace*{6.1em}
24.38\,dB (0.74) \hspace*{6.1em}
28.31\,dB (0.82) \hspace*{6.1em}
\\[1em]
\includegraphics[width=.1\textwidth]{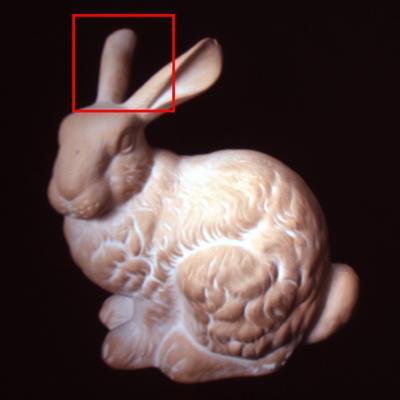}
\includegraphics[width=.1\textwidth]{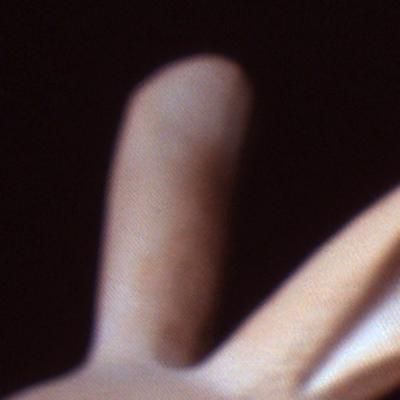}
\quad
\includegraphics[width=.1\textwidth]{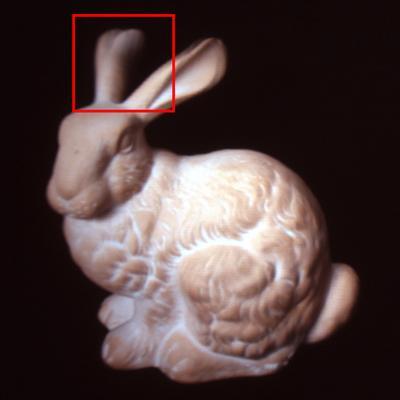}
\includegraphics[width=.1\textwidth]{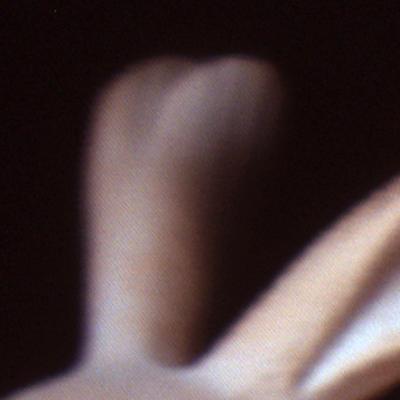}
\quad
\includegraphics[width=.1\textwidth]{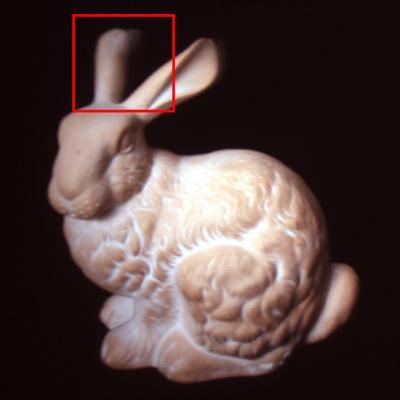}
\includegraphics[width=.1\textwidth]{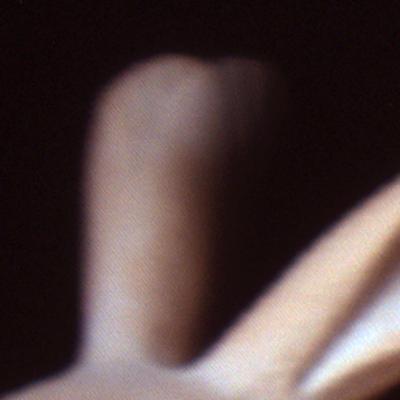}
\quad
\includegraphics[width=.1\textwidth]{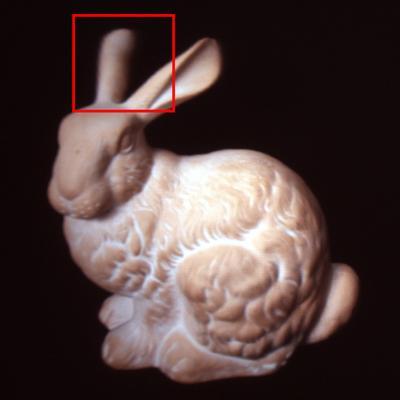}
\includegraphics[width=.1\textwidth]{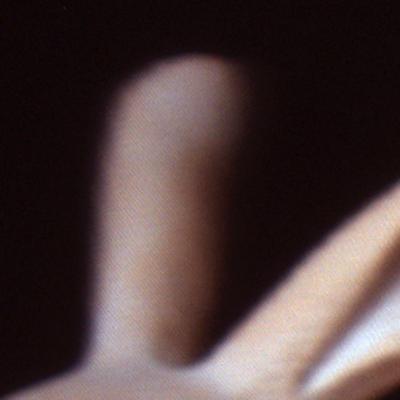}
\\
\hspace*{17.8em}
29.79\,dB (0.87) \hspace*{6.1em}
30.30\,dB (0.88) \hspace*{6.1em}
34.94\,dB (0.91) \hspace*{6.1em}
\\[1em]
\includegraphics[width=.1\textwidth]{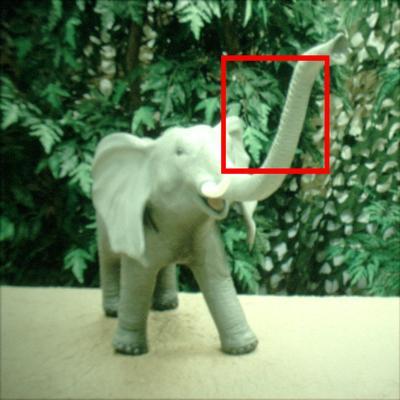}
\includegraphics[width=.1\textwidth]{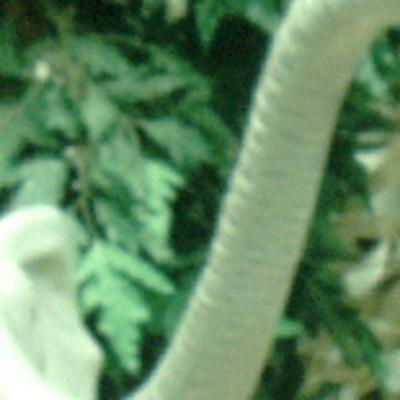}
\quad
\includegraphics[width=.1\textwidth]{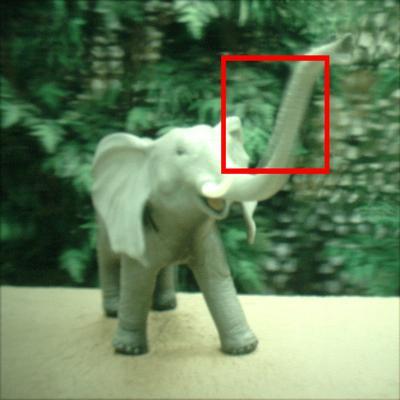}
\includegraphics[width=.1\textwidth]{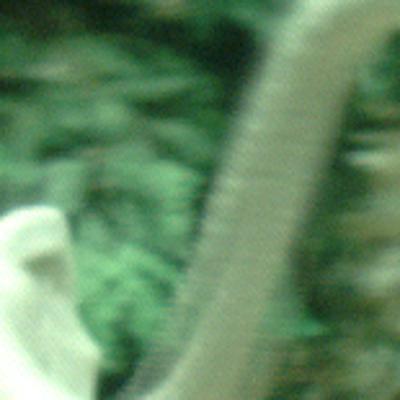}
\quad
\includegraphics[width=.1\textwidth]{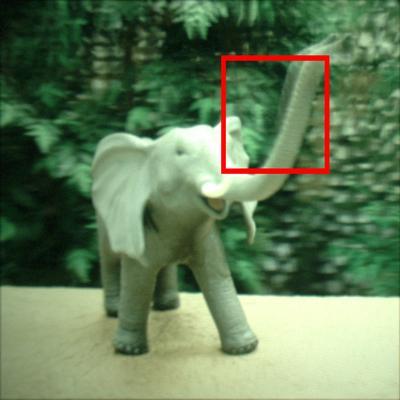}
\includegraphics[width=.1\textwidth]{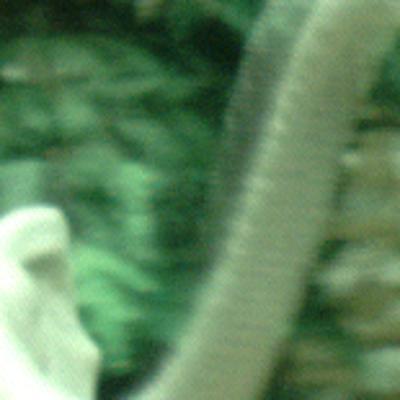}
\quad
\includegraphics[width=.1\textwidth]{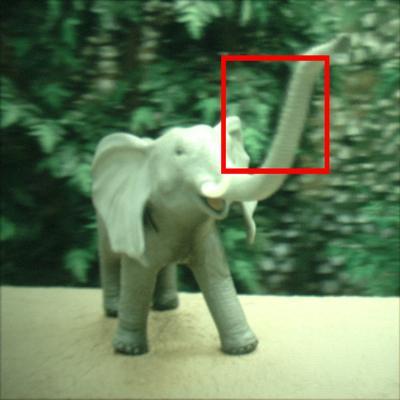}
\includegraphics[width=.1\textwidth]{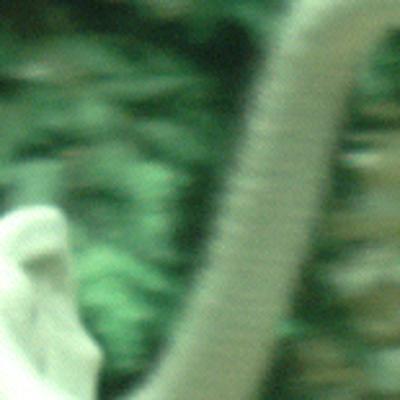}
\\
\hspace*{17.8em}
24.82\,dB (0.74) \hspace*{6.1em}
24.54\,dB (0.73) \hspace*{6.1em}
27.96\,dB (0.83) \hspace*{6.1em}
\\[1em]
\includegraphics[width=.1\textwidth]{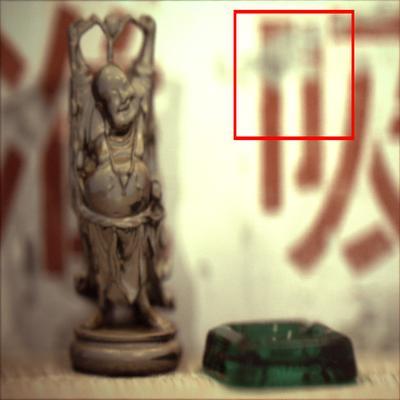}
\includegraphics[width=.1\textwidth]{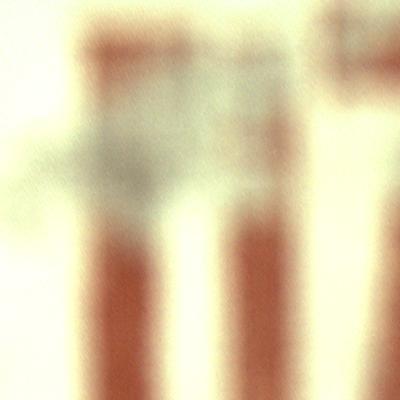}
\quad
\includegraphics[width=.1\textwidth]{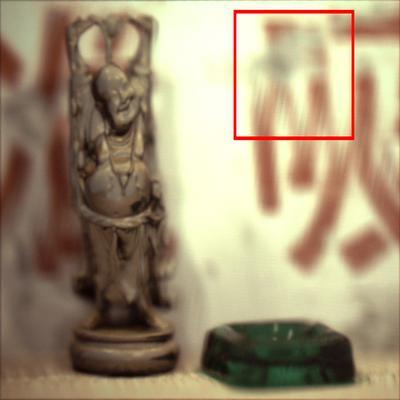}
\includegraphics[width=.1\textwidth]{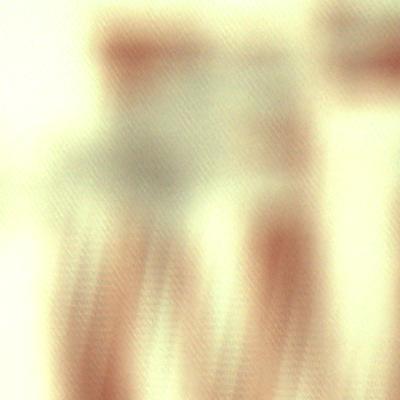}
\quad
\includegraphics[width=.1\textwidth]{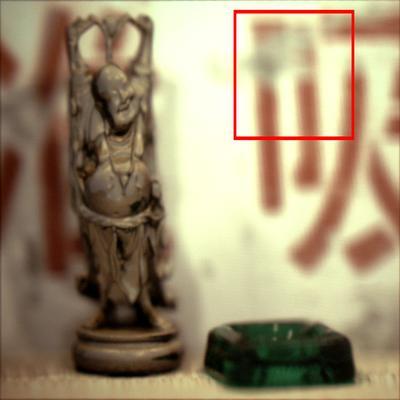}
\includegraphics[width=.1\textwidth]{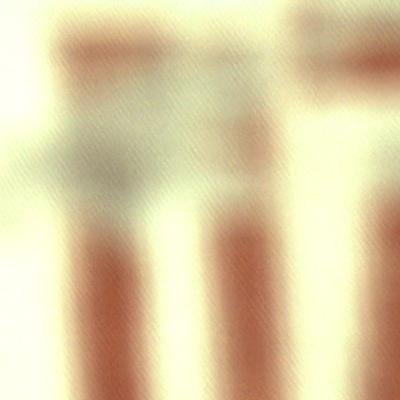}
\quad
\includegraphics[width=.1\textwidth]{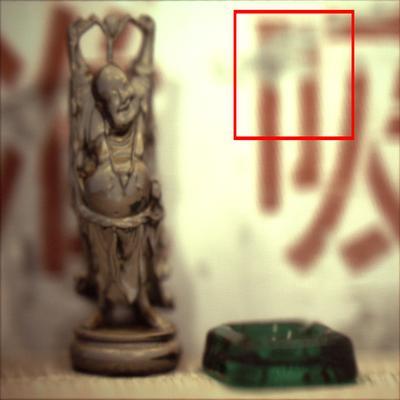}
\includegraphics[width=.1\textwidth]{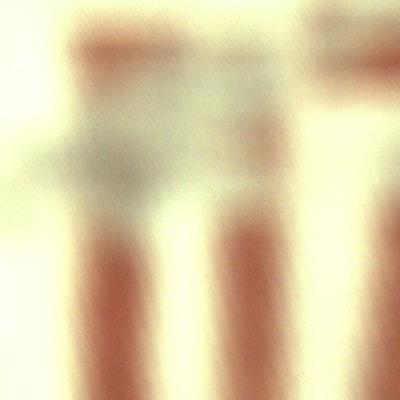}
\\
\hspace*{17.8em}
21.31\,dB (0.77) \hspace*{6.1em}
24.35\,dB (0.80) \hspace*{6.1em}
27.26\,dB (0.86) \hspace*{6.1em}
\\[1em]
\includegraphics[width=.1\textwidth]{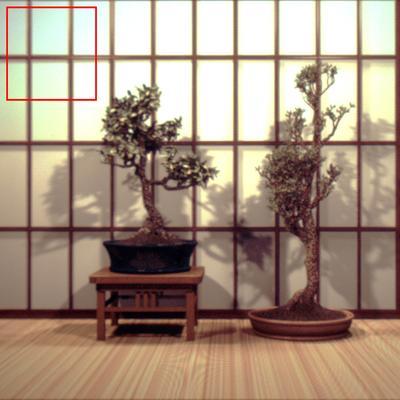}
\includegraphics[width=.1\textwidth]{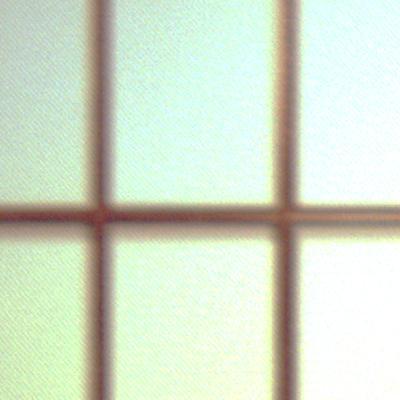}
\quad
\includegraphics[width=.1\textwidth]{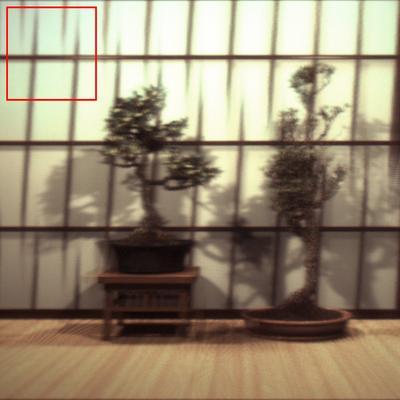}
\includegraphics[width=.1\textwidth]{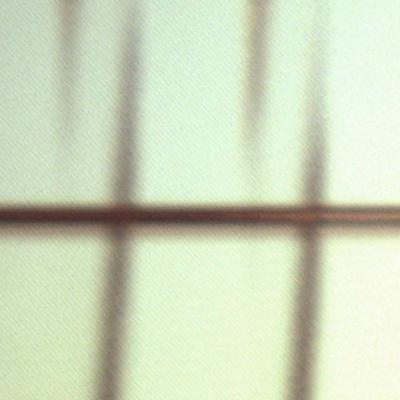}
\quad
\includegraphics[width=.1\textwidth]{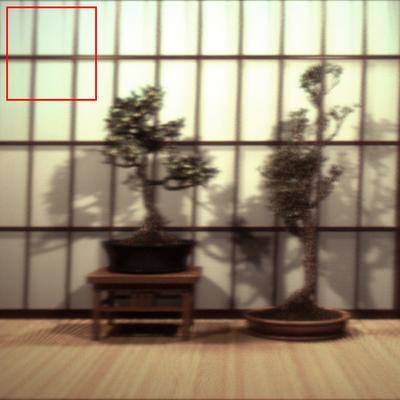}
\includegraphics[width=.1\textwidth]{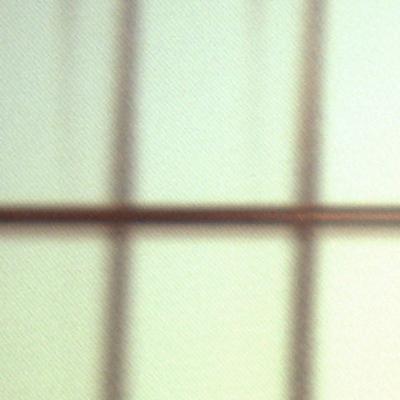}
\quad
\includegraphics[width=.1\textwidth]{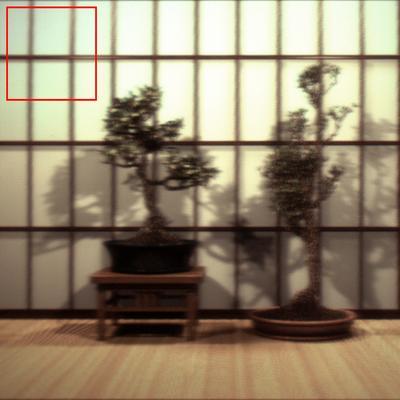}
\includegraphics[width=.1\textwidth]{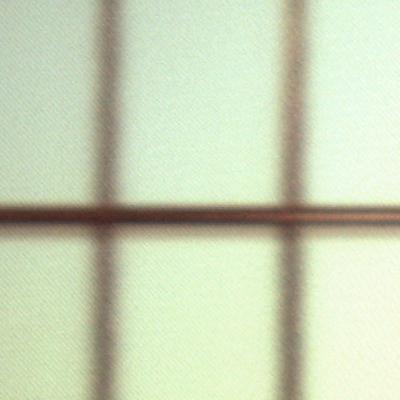}
\\
\hspace*{17.8em}
18.13\,dB (0.69) \hspace*{6.1em}
18.51\,dB (0.71) \hspace*{6.1em}
21.14\,dB (0.77) \hspace*{6.1em}
\\[-0.5em]
}
\subfigure[]{
\label{fig:real-lab-reference}
}
\hspace*{10.7em}
\subfigure[]{
\label{fig:real-lab-no-calib}
}
\hspace*{10.7em}
\subfigure[]{
\label{fig:real-lab-lee-ra}
}
\hspace*{10.7em}
\subfigure[]{
\label{fig:real-lab-proposed}
}
\caption{Example images observed on the 32\,in 3D display.
The images {\em Lego Knights}, {\em The Stanford Bunny},
{\em Elephant}, {\em Happy Buddha}, {\em Bonsai} (from top to bottom)
are rendered using different sets of parameters
and photographed nearly at the center position.
\subref{fig:real-lab-reference} Reference,
\subref{fig:real-lab-no-calib} No-calibration,
\subref{fig:real-lab-lee-ra} Calibration based on prior work \cite{lee2006image},
\subref{fig:real-lab-proposed} Calibration based on the proposed method.
The quality improvement by the proposed calibration is easily noticeable.
Two objective measures (PSNR in dB and SSIM index on a scale of 0 to 1) are
given below each image.
For better comparison, some regions are shown in magnification.
}
\label{fig:real-lab}
\end{figure*}

First, we calibrate a 32\,in ultra-high-definition (UHD)
display (with resolution 3\,840$\times$2\,160) 
that has a parallax barrier between the panel
and the back-light unit.
For the camera, we use Point Grey's Flea$^\text{\textregistered}$3 of 
the resolution 4\,096$\times$2\,160.
We take two photos, each nearly at $d_1\approx\text{700\,mm}$ and 
at $d_2\approx\text{1\,000\,mm}$,
respectively.
The calibration is conducted on a PC workstation
equipped with Intel Xeon E5-1660 processor and
16GB memory.
Given the input images, the entire process 
has only taken as little as 1.99\,s.
Fig.~\ref{fig:real-lab} exhibits 
the visual quality improvement by calibration
for five example images.

The images in Fig.~\ref{fig:real-lab} have been photographed
nearly at the center position
and nearly at the optimal viewing distance,
while the 3D rendering is performed according to
different sets of display parameters --
\subref{fig:real-lab-no-calib} with the designed parameters;
\subref{fig:real-lab-lee-ra} with the parameters calibrated
based on prior work \cite{lee2006image};
\subref{fig:real-lab-proposed} with the parameters
calibrated based on the proposed method.
In Fig.~\ref{fig:real-lab-no-calib},
the edges in wall bricks, rabbit's ear, elephant's nose,
and door frames appear as if multiply-exposed, and 
the Chinese character on the background of {\em Happy Buddha} suffers
from severe geometric distortion.
In Fig.~\ref{fig:real-lab-lee-ra},
the distortions have been much alleviated 
but still remain.
They almost disappear in Fig.~\ref{fig:real-lab-proposed}.
Fig.~\ref{fig:real-lab} also shows
the visual quality metrics in terms of 
peak-signal-to-noise-ratio (PSNR) 
as well as structural similarity (SSIM) index \cite{wbss04}.
Both measures require a reference image for the quality assessment.
For the reference,
we must be able to observe the misalignment-artifact-free image
through the optical elements.
To do this, we carefully identify the view $\gamma$ corresponding
to the camera position and subsequently display the viewpoint image
without giving any disparity.
Then, what the observer will actually get is a 2D image,
not a 3D image, but nevertheless it delivers 
the correct target contents
(in reduced resolution but without misalignment effects)
for the specific viewpoint where the camera is located.
The proposed scheme consistently
shows higher scores, in both objective criteria, 
than the others
(by 4--7\,dB in PSNR, 0.06--0.1 in SSIM index on average).
An interesting point is that, in this case,
the maximum difference among the parameters is quite small
-- smaller than 1\,$\mu$m for pitch,
smaller than 0.1\,$^\circ$ for slanted angle.
But as manifested in the figure,
the visual quality varies a lot,
indicating why accurate calibration is vital to 3D displays.

\begin{figure*}[t]
\centering
\subfigure[]{
\includegraphics[width=.22\textwidth]{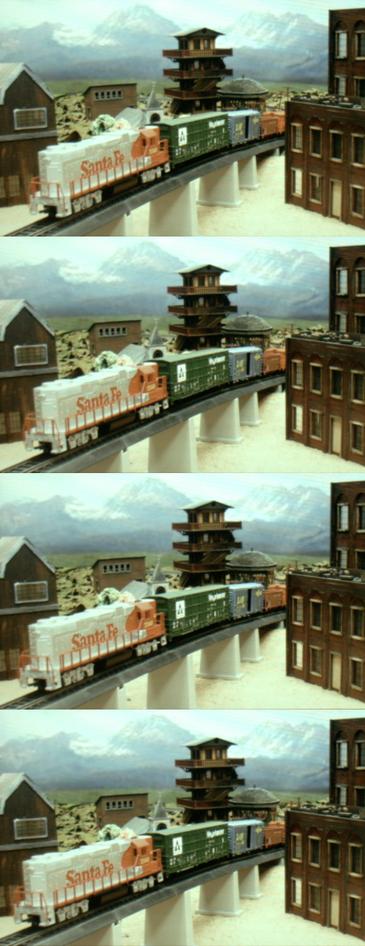}
\label{fig:parallax-lab-reference}
}
\subfigure[]{
\includegraphics[width=.22\textwidth]{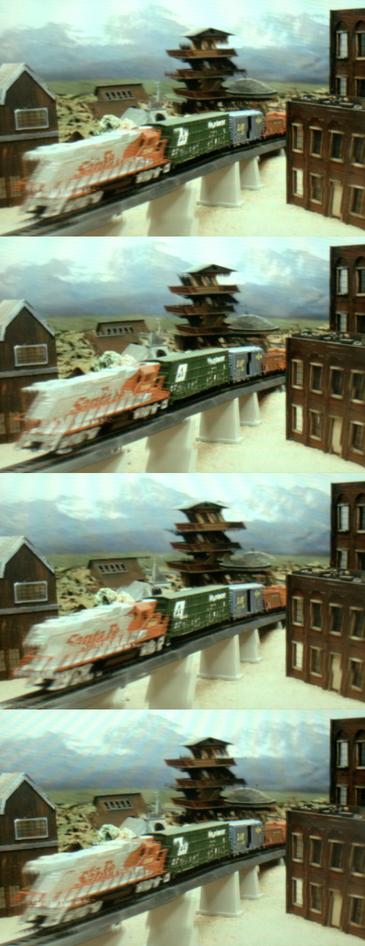}
\label{fig:parallax-lab-no-calib}
}
\subfigure[]{
\includegraphics[width=.22\textwidth]{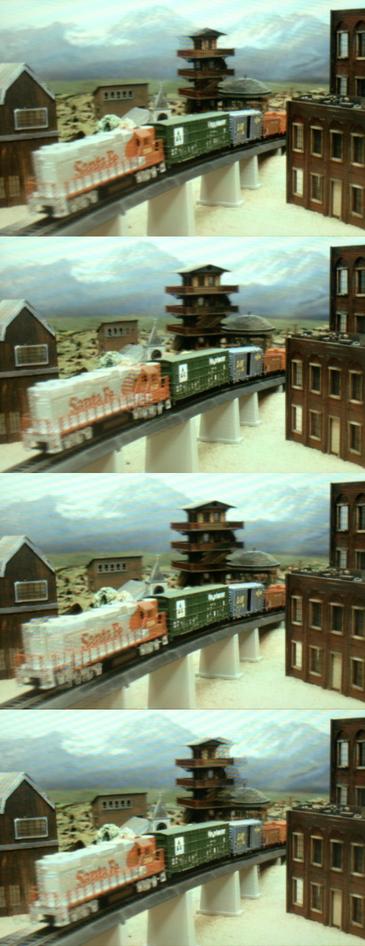}
\label{fig:parallax-lab-lee-ra}
}
\subfigure[]{
\includegraphics[width=.22\textwidth]{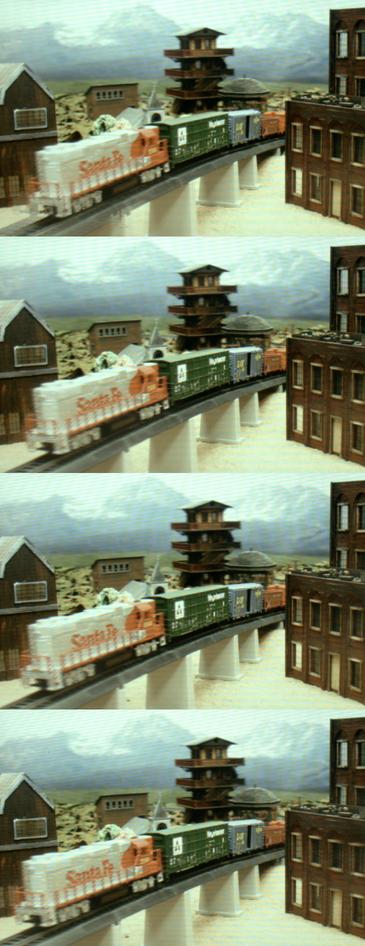}
\label{fig:parallax-lab-proposed}
}
\caption{Example images observed at four different view points.
The image {\em Train} is rendered using different sets of
parameters.
\subref{fig:parallax-lab-reference} Reference,
\subref{fig:parallax-lab-no-calib} No-calibration,
\subref{fig:parallax-lab-lee-ra} Calibration based on prior work \cite{lee2006image},
\subref{fig:parallax-lab-proposed} Calibration based on the proposed method.
The motion parallax is clearly observed in \subref{fig:parallax-lab-proposed}.
The train, buildings, and mountains move by different amounts
and/or in different directions as the view point changes.}
\label{fig:parallax-lab}
\end{figure*}

Next, we calibrate a tablet PC
with 2\,560$\times$1\,600 pixels on its 10\,in panel.
\begin{figure}[t]
\centering
\subfigure[]{
\includegraphics[width=.22\textwidth]{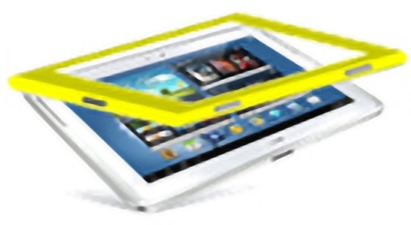}
\label{fig:tablet}
}
\subfigure[]{
\includegraphics[width=.22\textwidth]{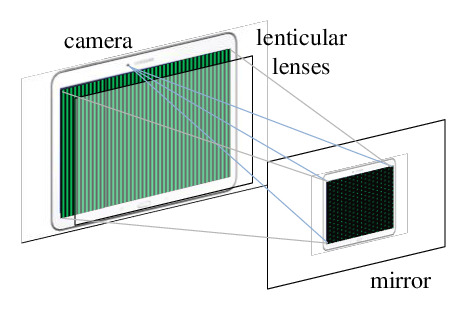}
\label{fig:tablet-calib}
}
\caption{Mobile 3D display calibration using on-board camera. 
\subref{fig:tablet} 3D display implemented on a tablet PC.
\subref{fig:tablet-calib} Calibration using the on-board camera and a mirror.
The tablet has a detachable lenticular cover
for 2D/3D switching. With the cover closed,
the tablet behaves like a 3D display.
The calibration can be performed with the on-board camera.
A pattern image is displayed on the panel,
and then a couple of images
reflected by the mirror are captured with the camera.
Each time, the mirror must be at different positions.
Except that the images need to be flipped horizontally,
the calibration procedure remains the same. 
}
\label{fig:tablet-concept}
\end{figure}
A detachable lenticular sheet serves as the frontal cover
of the tablet (see Fig.~\ref{fig:tablet}).
With the cover flipped open,
the tablet normally behaves as a 2D display,
but it turns into a 3D display
when the cover is closed.
This kind of design enables 2D/3D switching at little cost.
But, on each flip,
the lenticular sheet is likely to move
from where it was previously,
demanding a new instantaneous calibration.
We can use the built-in camera of the tablet PC,
together with a mirror,
for the calibration,
as conceptually depicted in Fig.~\ref{fig:tablet-calib}.
Except that the captured images need to be flipped horizontally,
the calibration procedure remains the same.
\begin{figure*}[t]
\centering
{\small
\includegraphics[width=.19\textwidth]{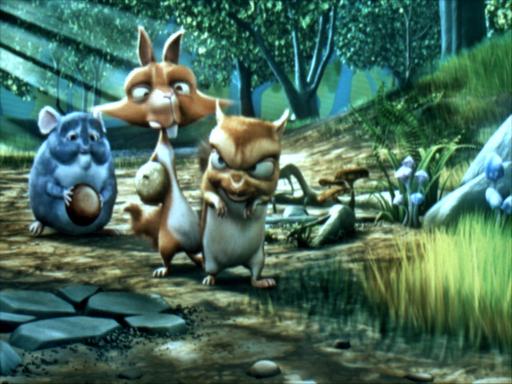}
\quad
\includegraphics[width=.19\textwidth]{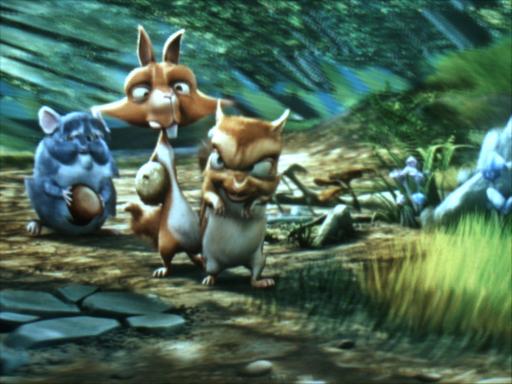}
\quad
\includegraphics[width=.19\textwidth]{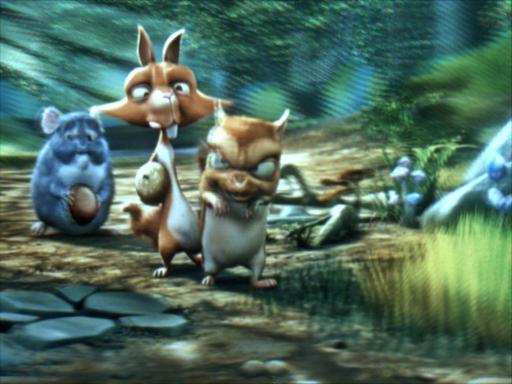}
\quad
\includegraphics[width=.19\textwidth]{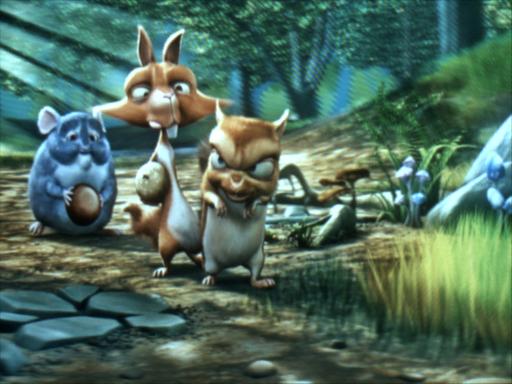}
\\
\hspace*{18.8em}
20.09\,dB (0.66) \hspace*{5.9em}
21.04\,dB (0.68) \hspace*{5.9em}
25.68\,dB (0.80) \hspace*{5.9em}
\\[1em]
\includegraphics[width=.19\textwidth]{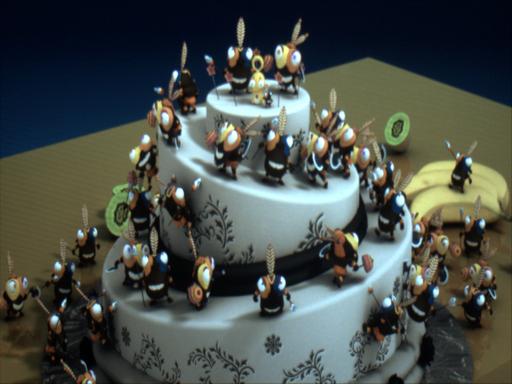}
\quad
\includegraphics[width=.19\textwidth]{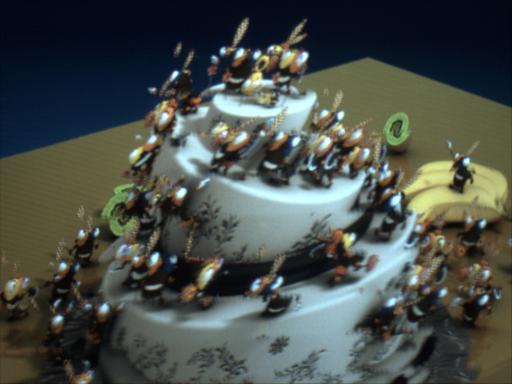}
\quad
\includegraphics[width=.19\textwidth]{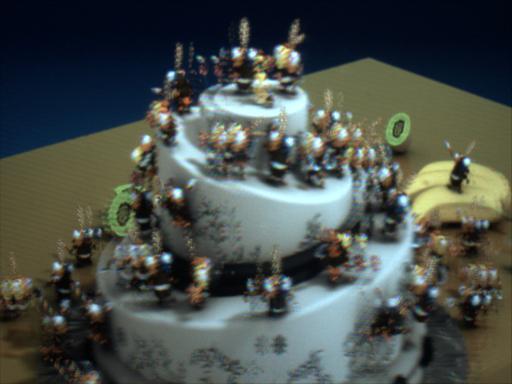}
\quad
\includegraphics[width=.19\textwidth]{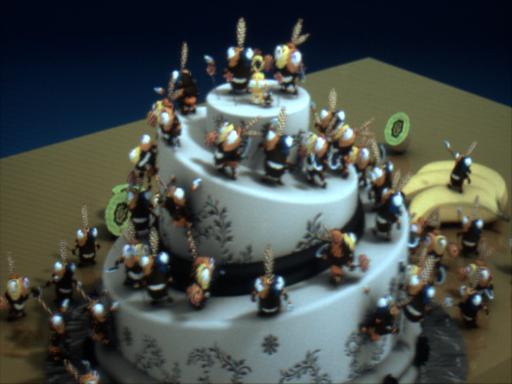}
\\
\hspace*{18.8em}
19.72\,dB (0.80) \hspace*{5.9em}
20.93\,dB (0.80) \hspace*{5.9em}
24.91\,dB (0.87) \hspace*{5.9em}
\\[1em]
\includegraphics[width=.19\textwidth]{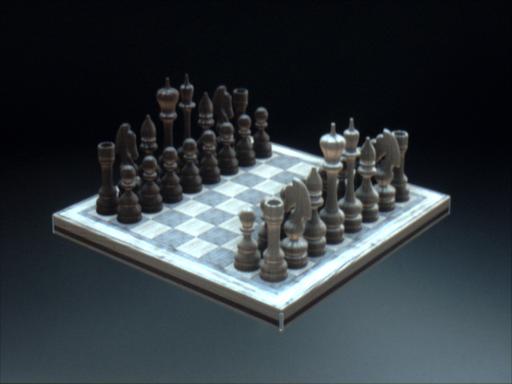}
\quad
\includegraphics[width=.19\textwidth]{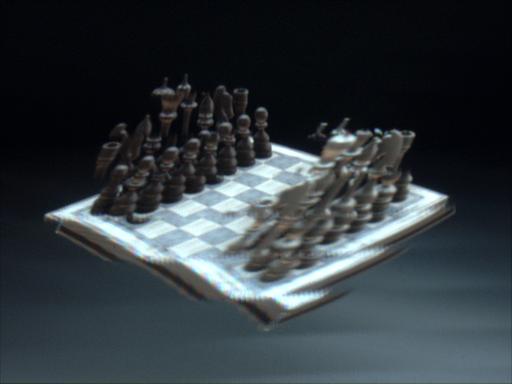}
\quad
\includegraphics[width=.19\textwidth]{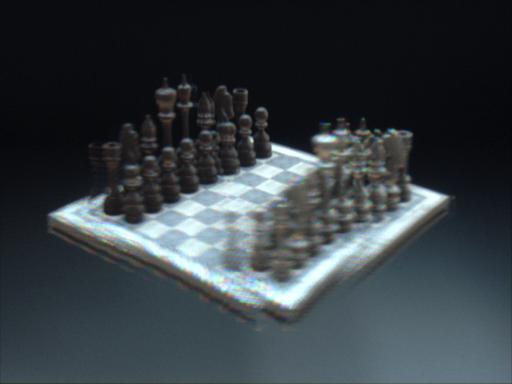}
\quad
\includegraphics[width=.19\textwidth]{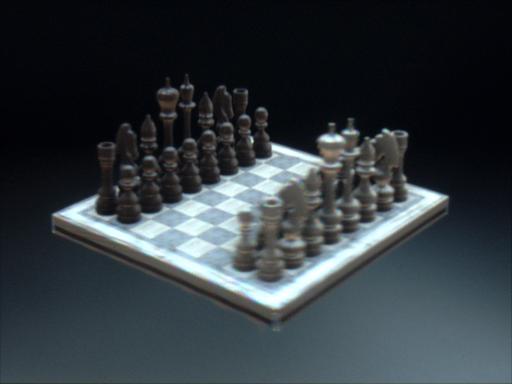}
\\
\hspace*{18.8em}
23.60\,dB (0.91) \hspace*{5.9em}
25.06\,dB (0.92) \hspace*{5.9em}
29.58\,dB (0.95) \hspace*{5.9em}
\\[1em]
\includegraphics[width=.19\textwidth]{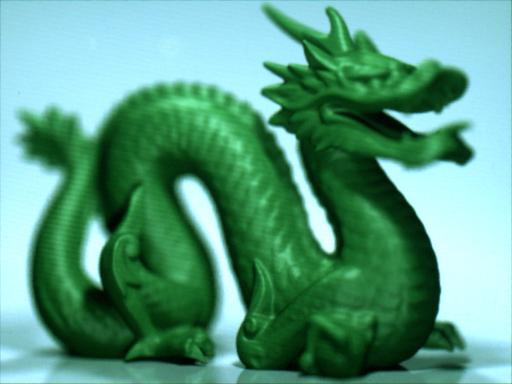}
\quad
\includegraphics[width=.19\textwidth]{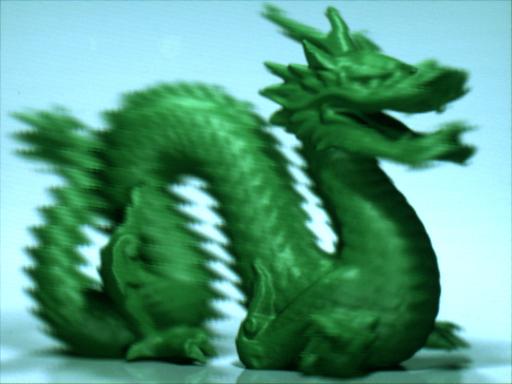}
\quad
\includegraphics[width=.19\textwidth]{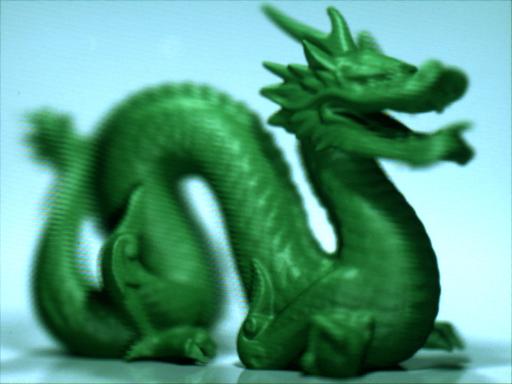}
\quad
\includegraphics[width=.19\textwidth]{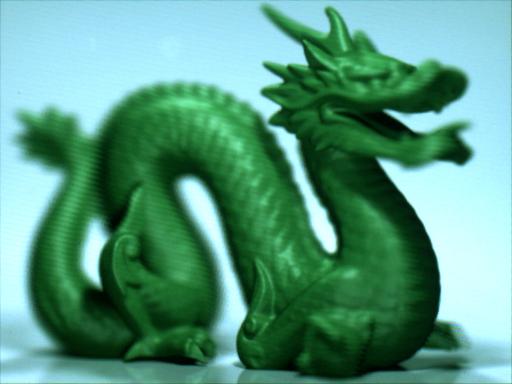}
\\
\hspace*{18.8em}
24.43\,dB (0.88) \hspace*{5.9em}
23.88\,dB (0.88) \hspace*{5.9em}
32.81\,dB (0.89) \hspace*{5.9em}
\\[1em]
\includegraphics[width=.19\textwidth]{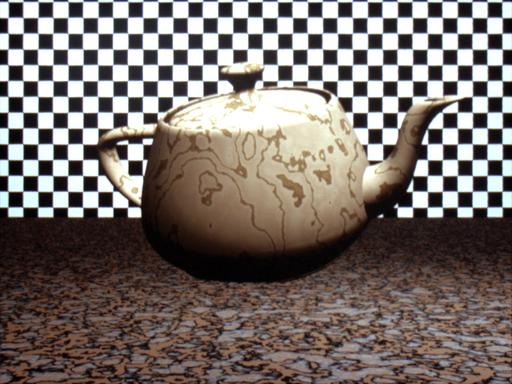}
\quad
\includegraphics[width=.19\textwidth]{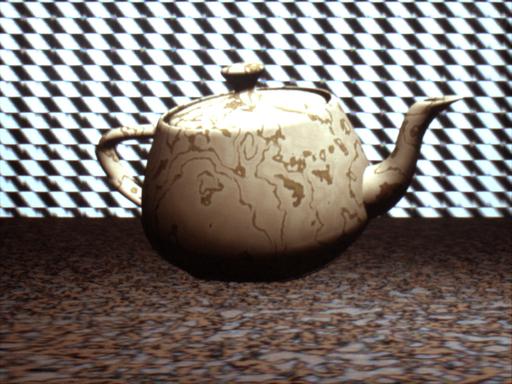}
\quad
\includegraphics[width=.19\textwidth]{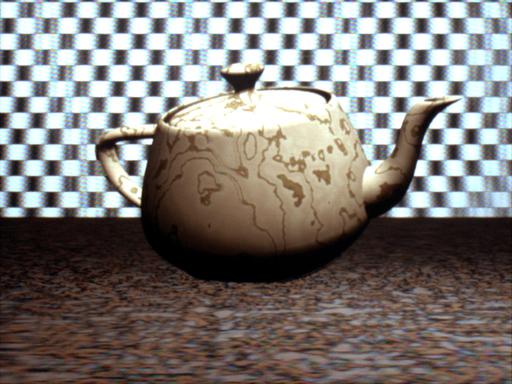}
\quad
\includegraphics[width=.19\textwidth]{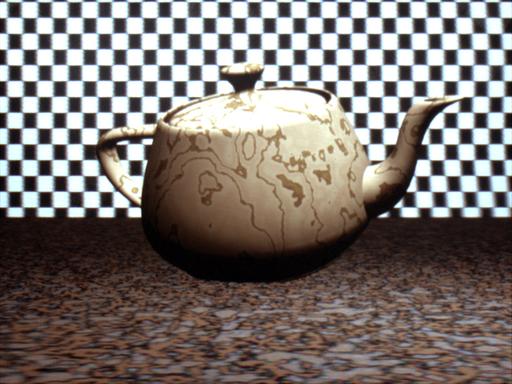}
\\
\hspace*{18.8em}
10.29\,dB (0.58) \hspace*{5.9em}
10.84\,dB (0.59) \hspace*{5.9em}
20.53\,dB (0.82) \hspace*{5.9em}
\\[-0.5em]
}
\subfigure[]{
\label{fig:tablet-lab-reference}
}
\hspace*{10.5em}
\subfigure[]{
\label{fig:tablet-lab-no-calib}
}
\hspace*{10.5em}
\subfigure[]{
\label{fig:tablet-lab-lee-ra}
}
\hspace*{10.5em}
\subfigure[]{
\label{fig:tablet-lab-proposed}
}
\caption{Example images observed on the 10\,in mobile 3D display.
The images {\em Big Buck Bunny}, {\em Cake}, 
{\em Chess}, {\em Green Dragon}, {\em Teapot}
(from top to bottom) are rendered using different sets of parameters
and photographed nearly at the center position.
\subref{fig:tablet-lab-reference} Reference,
\subref{fig:tablet-lab-no-calib} No-calibration,
\subref{fig:tablet-lab-lee-ra} Calibration based on prior work \cite{lee2006image},
\subref{fig:tablet-lab-proposed} Calibration based on the proposed method.
The quality improvement by the proposed calibration is clearly noticeable,
particularly along edges and in the objects with large binocular disparities.
Two objective measures (PSNR in dB and SSIM index on a scale of 0 to 1) are
given below each image.
Notice that more distortions happen without calibration, in comparison with Fig.~\ref{fig:real-lab}, which is due to a larger amount of optical misalignment. However, when calibrated using the proposed scheme, the resulting images have good visual quality nearly equal to those in Fig.~\ref{fig:real-lab}.  
}
\label{fig:tablet-lab}
\end{figure*}
Fig.~\ref{fig:tablet-lab} shows four example images
rendered with the calibrated display parameters.
The image distortion has remarkably decreased
in comparison with two other schemes: (i) no-calibration,
(ii) calibration based on prior work \cite{lee2006image}.
Fig.~\ref{fig:parallax-lab} shows
example images photographed at multiple view points
to verify the motion parallax along the horizontal direction.
We see that, in the figure,
the recognizability of the motion parallax is also enhanced
quite much by fixing the rendering parameters.

\section{Conclusion}
\label{sec:concl}

Despite the recent advancement,
3D displays have faced several issues to solve,
regarding the visual quality, for active proliferation. 
Some of them are fundamental,
unavoidable even if the display behaves without error
(e.g., resolution reduction, light leakage). But others are not,
while being as important issues.
In this paper, we dealt with one such matter --
correcting image distortions
caused by optical misalignment.

First, we established an observation model
when the observer looks at images through the optical elements
of the 3D display.
We also considered what the observation would be like
if misalignment happens.
Then, we proposed a 3D display calibration 
which decodes the correct display parameters
from the observation.
Given two photos
of a pattern image displayed on the 3D panel,
each taken at different positions, 
the analysis is conducted in frequency domain.
All procedures are fully automated
and the computation time spent on estimating
all display parameters (i.e., pitch, slanted angle, gap, and offset)
is less than two seconds.
The efficiency of the proposed method makes it
applicable on-the-fly whenever it is necessary.
In a set of experiments with synthetic dataset,
the proposed calibration method showed quite high accuracy, 
with the estimation error
0.0031\,$^\circ$ for slanted angle, 0.02\,$\mu$m for pitch, 
74\,$\mu$m for offset, 13\,$\mu$m for gap on average, 
a half order of magnitude higher than prior work.
With real-life displays, 
the proposed method has also demonstrated a significant improvement of
visual quality of the observed images.

\bibliographystyle{IEEEtran}
\bibliography{dispcalib}

\end{document}